\documentclass[conference]{ieee_tran}
\IEEEoverridecommandlockouts  
\usepackage{times}

\usepackage[sort&compress,numbers]{natbib}
\usepackage{multicol}
\usepackage[bookmarks=true]{hyperref}
\usepackage[table]{xcolor}
\usepackage{makecell}
\usepackage{amsmath,amssymb,amsfonts}
\usepackage{multirow}
\usepackage{graphicx}
\usepackage{arydshln}
\usepackage{color,soul}
\usepackage{xcolor}
\usepackage{pifont}
\usepackage{fancyhdr}
\begin{document}

\title{World Models for General Surgical Grasping\\

}



\author{Hongbin Lin$^{1}$, Bin Li$^{1}$, Chun Wai Wong$^{1}$, Juan Rojas$^{2}$, Xiangyu Chu$^{1}$, and Kwok Wai Samuel Au$^{1}$
\\
$^{1}$The Chinese University of Hong Kong \qquad $^{2}$Lipscomb University \\
 \tt\small \{hongbinlin.link,binli.link,xiangyuchu,samuelau\}@cuhk.edu.hk \\ \tt\small juan.rojas@lipscomb.edu, brianwongcw@gmail.com
} 

\maketitle

\fancypagestyle{firstpage}
{
    \renewcommand{\headrulewidth}{0pt}
    \fancyhead[L]{\textbf{\textsf{Robotics: Science and Systems 2024 \\ Delft, Netherlands, July 15-July 19, 2024}}}    
    \fancyhead[R]{}
    \fancyhead[C]{}
    \fancyfoot[]{}
}
\thispagestyle{firstpage}
\begin{abstract}
\color{black}
   Intelligent vision control systems for surgical robots should adapt to unknown and diverse objects while being robust to system disturbances. Previous methods did not meet these requirements due to mainly relying on pose estimation and feature tracking. We propose a world-model-based deep reinforcement learning framework ``Grasp Anything for Surgery'' (GAS), that learns a pixel-level visuomotor policy for surgical grasping, enhancing both generality and robustness. In particular, a novel method is proposed to estimate the values and uncertainties of depth pixels for a rigid-link object's inaccurate region based on the empirical prior of the object's size; both depth and mask images of task objects are encoded to a single compact 3-channel image (size: 64x64x3) by dynamically zooming in the mask regions, minimizing the information loss. The learned controller's effectiveness is extensively evaluated in simulation and in a real robot. Our learned visuomotor policy handles: i) unseen objects, including 5 types of target grasping objects and a robot gripper, in unstructured real-world surgery environments, and ii) disturbances in perception and control. 
   Note that we are the first work to achieve a unified surgical control system that grasps diverse surgical objects using different robot grippers on real robots in complex surgery scenes (average success rate: $69\%$). Our system also demonstrates significant robustness across 6 conditions including background variation, target disturbance, camera pose variation, kinematic control error, image noise, and re-grasping after the gripped target object drops from the gripper. Videos and codes can be found on our project page: \url{https://linhongbin.github.io/gas/}.
\color{black}
\end{abstract}

\IEEEpeerreviewmaketitle
\section{Introduction}

Automating tedious and repetitive grasping tasks in robot-assisted surgery (RAS) can significantly relieve physical and mental fatigue in surgeons. Frequent surgical tasks include the grasping of objects like needles, gauze, sponges, thread, and processes like debridement. 

Traditional solutions first estimate the Cartesian poses of task-related objects from observed camera images and then manipulate goal objects  \cite{d2018automated, lu2020dual, ozguner2021visually,schwaner2021autonomous,xu2021surrol,chiu2021bimanual,bendikas2023learning,long2023human,huang2023demonstration,sen2016automating,sundaresan2019automated,wilcox2022learning,joglekar2023suture,lu2019surgical,kehoe2014autonomous,seita2018fast,fan2024learn,hwang2022automating}. Therefore, these methods are not robust against pose estimation error and will lead to grasping failures, or even catastrophic results. In  \cite{zhong2019dual}, the system directly tracks the feature coordinates of the observed images to increase visual robustness, instead of object pose estimation. However, both pose-estimation and feature-tracking methods require full geometric models or assume geometric shapes to infer an arbitrary point position's information. As such, it is difficult to adapt online to unseen objects in unstructured environments. To bridge the gap between research and real surgery applications, we aim to enhance 
i) generalization of unseen objects in unstructured environments, and ii) the robot's robustness across disturbances in perception and control.

 \begin{figure}[!tbp]
  \centering
  \includegraphics[width=\hsize]{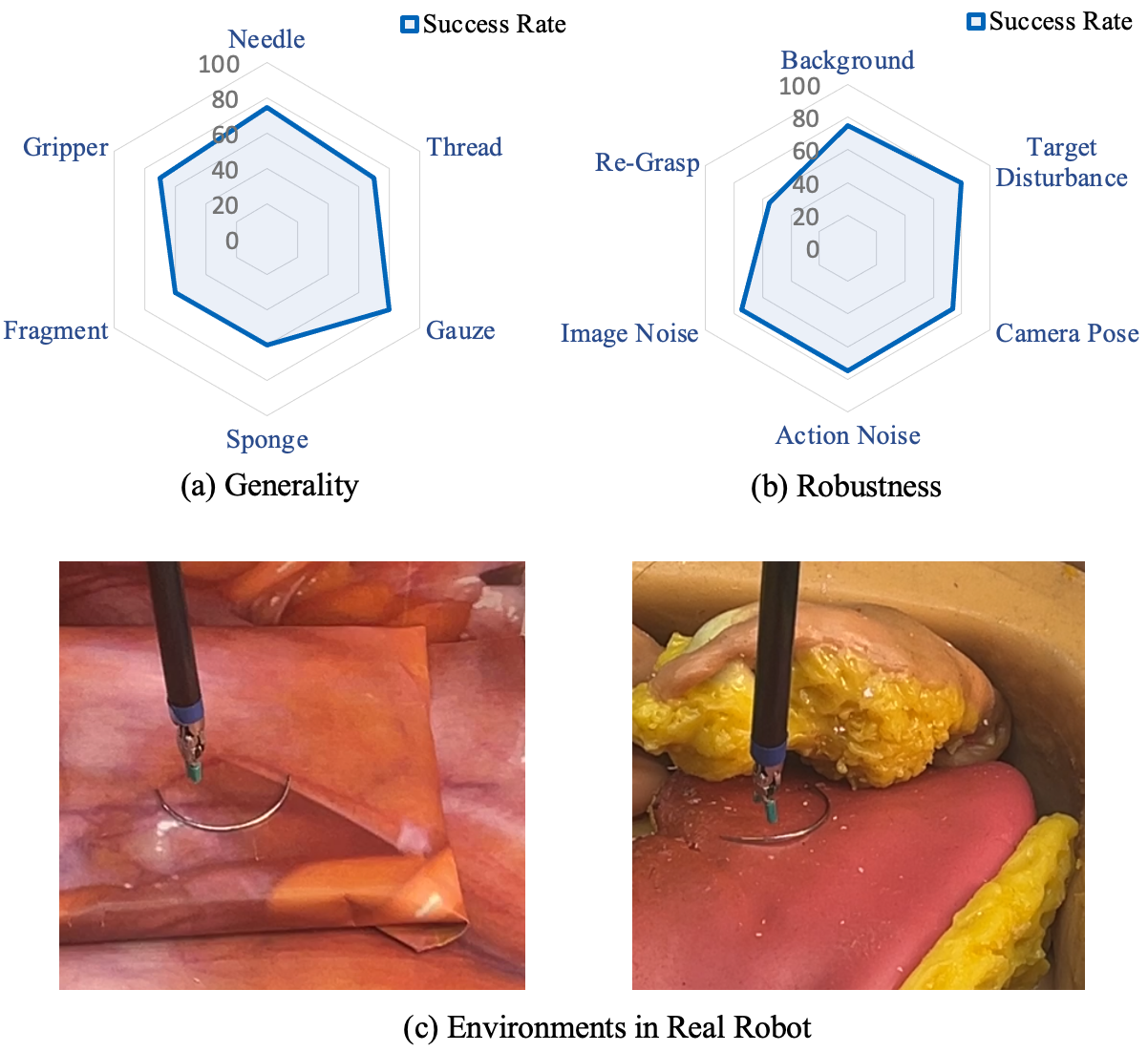}
  \caption{Generality, robustness, and evaluated environments of our visuomotor controller. (a) The success rates of our visuomotor controller for unseen objects, including 5 types of grasping objects and a robot gripper, are demonstrated. (b) We also show the controller's robustness across 6 conditions including background variation, target disturbance, camera pose variation, kinematic control error, image noise, and re-grasping after the gripped target object drops from the gripper. (c) A robot gripper is actuated by the controller to grasp a needle on a rectangle phantom (left bottom) and a liver phantom (right bottom).}
  \vspace{-0.45cm}
  \label{fig1}
\end{figure}

\color{black}Visuomotor control learning has been proven to improve performance in the aforementioned core factors: generalization of unseen objects \cite{kalashnikov2018qt}, and robustness to perception disturbance \cite{seo2023multi,kalashnikov2018qt}. \color{black} Recent advances in \textit{world models} \cite{ha2018world}, a model-based visuomotor learning method, show state-of-the-art (SOTA) performance in robot grasping \cite{wu2023daydreamer,seo2023multi}. World models learn to encode the spatial and temporal representation of the environment in an unsupervised manner using auto-encoding \cite{kingma2013auto} and recurrent neural networks \cite{hafner2019learning}, respectively. Given learned world models, visuomotor policies used the encoded featured as input and can be learned by reinforcement learning (RL) \cite{hafner2020mastering}, imitation learning \cite{lancaster2023modem},  and planning \cite{hafner2019learning}. 
However, deploying the SOTA world-model-based learning for general surgical grasping is more challenging compared to general robotic grasping.
First, the area of projected pixels for centimeter-level task objects (i.e. visual signals) is extremely small w.r.t. that of the background (i.e. visual noise) in a RAS canonical camera setting, typically exhibiting signal-to-noise ratios of 1:99. Thus, visual observations with low signal-to-noise ratios in surgical grasping make auto-encoding learning for world models challenging. Second, visuomotor control in surgical grasping demands millimeter-level accuracy of depth images. Yet, conventional depth cameras hardly meet such a requirement due to the limitation of their sensing principles. For instance, the Intel Realsense D435 camera (used in this work), uses a structured-light depth camera that captures the projected infrared pattern via stereo cameras and measures depth using triangulation \cite{haider2022can}. With this camera, the depth error for a task-related object (e.g., the gripper tip) is prone to be large since the light pattern can not be fully projected onto the object's surface due to its small and narrow shape. The large error introduces uncertainty rendering the visual task more challenging. \color{black}

\color{black}In this paper, we propose a unified framework, \textbf{G}rasp \textbf{A}nything for \textbf{S}urgery (GAS), for general surgical grasping. Our work is the first to learn a visuomotor policy that can pick up unseen diverse surgical objects including needles, gauzes, sponges, and threads as well as perform debridement with different robot grippers in complex RAS environments (see Fig. \ref{fig1}c). Our work achieves strong performance with an average of 69 $\%$ success rate in our generalization study (see Fig. \ref{fig1}a). We also demonstrate significant robustness across 6 conditions including background variation, target disturbance, camera pose variation, kinematic control error, image noise, and re-grasping after the gripped target object drops from the gripper, without significant performance degradation (see Fig. \ref{fig1}b)\color{black}. In summary, our primary contributions are:
\color{black}
\begin{itemize}
   \item Proposal of the first unified framework that grasps diverse unseen surgical objects with different unseen grippers in realistic real-world environments of RAS. Our visuomotor policy, which is trained in simulation and directly transferred to a real robot without further finetuning, demonstrates high generality with significant performance (average 69\% success rate) and extraordinary robustness against disturbances in the environment.

   \item Proposal of a novel method to estimate the values and uncertainties of depth pixels for a rigid-link object's inaccurate region based on the empirical prior of the object's size. We estimate the imprecise depth pixels of task objects captured by an Intel Realsense D435 RGB-D camera and demonstrate that visuomotor learning can benefit from our depth estimation.
   
   \item Extension of Dynamic Spotlight Adaptation \cite{lin2023end}, a compact visual representation, to the setting of general surgical grasping. The proposed visual representation encodes both depth and mask images of task objects (size: 600x600) into a single compact 3-channel image (size: 64x64x3) by dynamically zooming in the mask regions, minimizing the loss of mask coordinates information as well as the quality of both depth and mask images. We demonstrate that our image representation can improve the performance of visuomotor learning.

\end{itemize}
\color{black}

\section{Related Work}
\subsection{Surgical Autonomous Grasping}

Previous researchers have studied how to grasp objects in surgical tasks including needle grasping, gauze picking, threads grasping, debridement, and peg transfer extensively. For the needle-grasping task, robots are required to either pick up a needle with an unknown pose or re-grasp a needle to hand over. A traditional way to perform needle grasping is first tracking the needle pose visually and then planning feasible trajectories based on the tracked pose via analytical solutions  \cite{d2018automated,schwaner2021autonomous} or sampled-based planning methods \cite{lu2020dual,ozguner2021visually}. Recent methods learned controllers via RL without the need for planning \cite{xu2021surrol,chiu2021bimanual,bendikas2023learning,long2023human,huang2023demonstration}. Planned trajectories are incorporated in the controller learning  \cite{chiu2021bimanual} and the grasping task is decomposed into subtasks  \cite{bendikas2023learning} to speed up the learning process. Some researchers explored easing the difficulty in pose estimation for needles by designing customized mechanical design  \cite{sen2016automating}, segmenting needle masks  \cite{sundaresan2019automated}, and changing needle pose to increase the visibility \cite{wilcox2022learning}. For other grasping objects, including threads  \cite{joglekar2023suture,lu2019surgical}, tissue fragments   \cite{kehoe2014autonomous,seita2018fast,fan2024learn} (in the debridement tasks) and blocks  \cite{hwang2022automating,xu2021surrol} (in the peg transfer task), Cartesian poses of the target objects are tracked and robots are controlled to grasp the objects via plan trajectories  \cite{kehoe2014autonomous,seita2018fast,hwang2022automating} or learned controller \cite{fan2024learn,xu2021surrol}. However, these methods are not robust to estimation error of tracked Cartesian poses and are hard to scale unseen objects due to the assumption on geometric model or shape. Some of these works explored the generality of their RL controllers to multiple grasping tasks  \cite{xu2021surrol,long2023human,huang2023demonstration}. Yet, they learned corresponding separate controllers for diverse grasping tasks and thus failed to learn similar control patterns across the tasks. Compared to their works, we learned a single visuomotor controller for diverse grasping tasks. Perhaps the work from Scheikl et al. \cite{scheikl2022sim} is the most similar to us, where a visuomotor policy is learned for tissue retraction in simulation using PPO \cite{schulman2017proximal}, a model-free RL, and unpaired image-to-image translation models are applied for sim-to-real visual domain adaptation. The learned visuomotor policy is further applied to real robots with $50\%$ success rate. However, they did not alleviate the limitations of pose tracking since the Cartesian position of the target grasping point and the gripper pose are needed to track visually for constructing their reward function. Furthermore, the initial pose of the target object is assumed to be fixed. Compared to their work, our visuomotor controller is learned by model-based RL without the need for pose estimation and performs a more challenging grasping task, where the robot is required to grasp an unseen target object with a random initial pose given a sparse-reward pose-tracking-free formulation.

\subsection{Visual Manipulation With World Models on Real Robots}
World models are learned with collected data from simulation \cite{kadi2023planet,mendonca2023alan,mandi2022cacti,hansen2022modem,lancaster2023modem,seo2023multi}, real robots \cite{wu2023daydreamer}, and human videos  \cite{mendonca2023structured}. Their control policies, trained by planning \cite{kadi2023planet,mendonca2023structured,mendonca2023alan}, imitation learning \cite{mandi2022cacti,hansen2022modem,lancaster2023modem}, and RL \cite{wu2023daydreamer,seo2023multi} based on the learned world models, are applied to visual manipulation tasks in real robots. The manipulated objects in these works can be categorized into rigid objects (e.g., door handles \cite{mendonca2023alan,mandi2022cacti}, plastic cups \cite{seo2023multi,lancaster2023modem}) and deformable objects (e.g., clothes \cite{kadi2023planet}, soft toy balls, sponge \cite{wu2023daydreamer}).  However, none of the previous works demonstrate their capacity to handle centimeter-level surgical objects (size: 10mm to 50mm) and visual depth uncertainty. Compared to their works, our learned visuomotor policy grasps centimeter-size surgical objects with strong performance and adapts to the uncertainty of observed depth images. In terms of system robustness, background noise is filtered out by a learned mask in \cite{mendonca2023alan}, and viewpoint randomization is applied to adapt viewpoint variations and disturbance \cite{seo2023multi}. Yet, the mask area for our task objects in the observed RGB-D images is much smaller compared to \cite{mendonca2023alan}. Furthermore, more comprehensive domain randomization, including camera pose, target disturbance, gripper disturbance, and image noise, is applied for visuomotor learning compared to \cite{seo2023multi}. 

 \begin{figure}[!tbp]
  \centering
  \includegraphics[width=0.8\hsize]{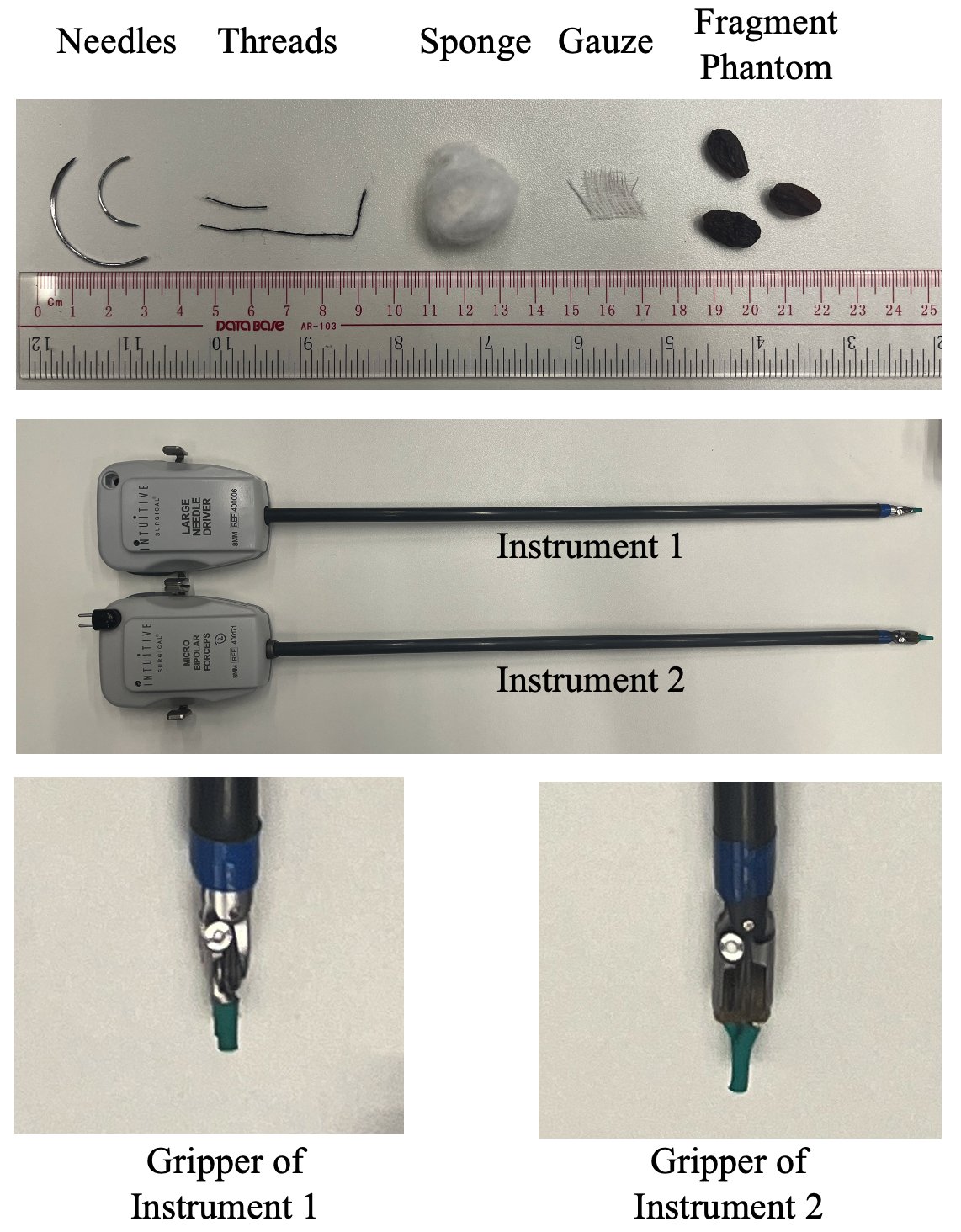}
  \caption{Diverse target objects and robot grippers in surgical grasping.}
  \vspace{-0.45cm}
  \label{fig:surgical_objects}
\end{figure}

\section{Problem Formulation and Background}

\subsection{Problem Setup and Assumptions}

\subsubsection{Problem Setup} We aim to actuate an unknown robot gripper to grasp and lift an unknown target object (10mm to 50mm) lying on a phantom plane via a learned visuomotor policy $\pi$ that maps 
temporal RGB-D images to a discrete action command. \color{black} Diverse target objects and robot grippers are utilized to evaluate the visuomotor policy, as shown in Fig. \ref{fig:surgical_objects}. \color{black} The initial poses of the target object 
and the robot gripper 
are set randomly within the workspace. At timestep $t$, a RGB-D image $I_t \in \mathbb{R}^{600\times600\times4}$ is observed, and a discrete command $a_t$ is predicted by the policy. Discrete commands actuate the gripper in 2mm increments across Cartesian dimensions and rotations steps of 10 degrees about the z-axis (normal to the plane) w.r.t to a world frame. Positive and negative directions are output independently and are mutually exclusive. When one direction is activated, the other will be null.
The 
gripper 
has an open and closed mode. 
As such, there are a total of 9 discrete commands $a_t$ at each timestep. 
\color{black}We set the horizon $H$ in our grasping tasks to 300 timesteps. The task is successful if the stuff is grasped and 
the target object is lifted 10 mm above the plane. The gripper and the target object are within a rectangular workspace during the grasping task. When the gripper's desired position is outside the workspace, the actuated position is clipped to the point in the workspace closest to the desired position.

\subsubsection{Assumptions} We assume that a) the robot arm will not exceed its joint limits; b) both target objects 
or the gripper can be partially but not fully occluded; and c) 
poses or visual feature coordinates of target objects or the gripper are not available. 

\subsection{POMDP Formulation}
We formulate the control problem of general surgical grasping as a discrete-time partially observable Markov Decision Process (POMDP). 
The latter consists of a 7-tuple $(S, A, R, T, O,\gamma, \Omega)$, where $S$ is a set of partially observable states; 
$A$ is a set of discrete actions; 
$R(s, a): S\times A \to \mathbb{R}$ is a reward function; 
$T$ is a set of conditional transition probabilities between states; 
$O$ is a set of conditional observation probabilities; 
$\gamma \in [0,1]$ is the discount factor; and
$\Omega$ is the observation. We aim to learn the control policy that maximizes its expected future discounted reward
$\mathbb{E}_{\pi}[ \sum_{i=t}^{T} \gamma^{i-t}r_{i}]$, where $r_i$ is the reward at time $i$.

\subsection{DreamerV2}
DreamerV2 \cite{hafner2020mastering} is a SOTA visual RL method that can learn visuomotor policy with high data efficiency and significant performance \cite{hafner2020mastering,wu2023daydreamer}. It consists of 2 components: i) world models, targeting to learn the POMDP dynamics with on-policy rollouts from environments, and ii) a control policy, targeting to optimize by merely simulated rollouts from the learned world models. Both world models and the control policy are learned simultaneously. In this paper, our proposed framework, GAS, is built upon DreamerV2 for visuomotor learning in a more challenging visual manipulation setting, surgical grasping. Details of DreamerV2 can be found in \cite{hafner2020mastering}.

 \begin{figure*}[!tbp]
  \centering
  \includegraphics[width=1.0\hsize]{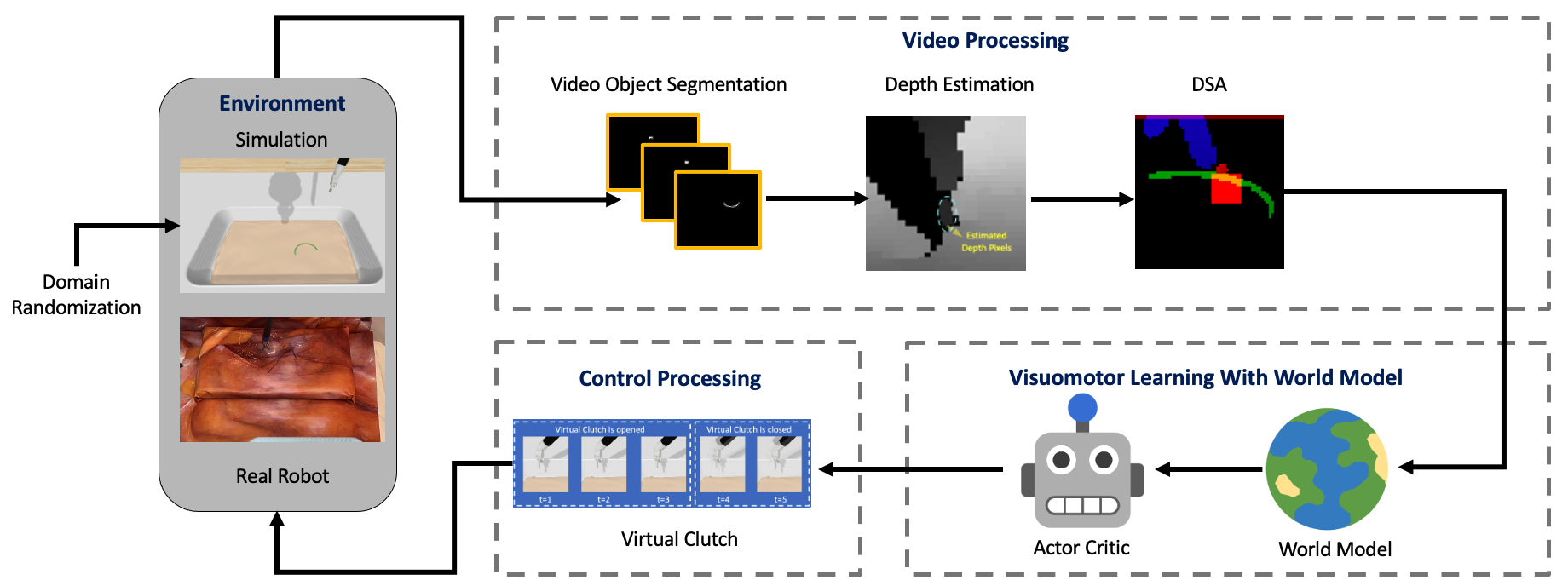}
  \caption{An overview of Grasp Anything for Surgery (GAS). Visual observations are processed by our proposed video processing, including video object segmentation, depth estimation, and Dynamic Spotlight Adaptation (DSA). A visuomotor policy, learned by world models, leverages the processed observations as the input. The predicted action of the policy, further processed by Virtual Clutch (VC), actuates the robot gripper to grasp target objects in simulation and a real robot. Furthermore, domain randomization is applied in simulation for visuomotor learning.} 
  \vspace{-0.45cm}
\label{fig:gas}
\end{figure*}

\section{Grasp Anything for Surgery}
\color{black}In this section, we will introduce our method, GAS, for visuomotor learning in general surgical grasping. We will first elaborate on our proposed visual processing, including video object segmentation, depth estimation, and visual representation, followed by our control processing, domain randomization as well as the design of rewards and terminations. Fig. \ref{fig:gas} shows the overview of our method\color{black}.

\color{black}
\subsection{Video Object Segmentation}

A key challenge in image-based visuomotor learning is how to reduce background interference and identify object classes from observed RGB-D images, especially in a complex surgery scene, where the area of projected pixels for both the gripper and the grasping object is extremely small w.r.t. that of the whole image (ratio: $1\%$). To alleviate the difficulty in visual learning of our visuomotor policy, each pixel of an RGB-D image is assigned a class and a corresponding class category. We formulate such a pixel-level assigning problem as a standard RGB video object segmentation (VOS) problem: given historical observed RGB images, we are required to compute binary masks $M=\{m_i\in\mathbb{R}^{600\times600}\}_{i=1}^K$ of $K$ task objects, aiming to assign the pixels of observed RGB-D image at time t with the class and the category. Table \ref{table:mask} shows the details of masks ($K=3$) in our grasping tasks.

Tracking object masks faces a series of challenges including target deformation, camera motion, etc \cite{yang2023track}. In the real-world scene of RAS, the motion blur \cite{allan20192017}, illumination variation such as shadows and specular reflections, and visual occlusions such as blood and camera lens fogging \cite{shvets2018automatic}, further exacerbate the issue. Furthermore, annotating a large number of video masks to facilitate data-driven methods of deep neural networks is tedious and time-consuming \cite{garcia2021image}.

We leverage a semi-VOS method, Track Anything Model (TAM) \cite{yang2023track}, to solve the issues of mask tracking in RAS. TAM combines the Segment-Anything Model (SAM) \cite{kirillov2023segment}, a large foundation model for image segmentation, and an advanced VOS model, XMem \cite{cheng2022xmem}. With a few clicks for annotation, TAM can track and segment any object in a given video with only one-pass inference. In this paper, we deploy TAM to the online object segmentation for visuomotor control, which can be divided into two stages:

\begin{itemize}
    \item \textit{Annotation Stage}: In this stage, we annotate the mask for the first frame of streaming RGB-D images. Users can label a fine mask for an object using SAM with only a few point prompts (around 2 clicks of point prompts for each object mask). The labeled image-mask pair is saved for further tracking.
    \item \textit{Online-Tracking Stage}: Given the image-mask pair of annotation in the first frame, object masks for the rest frames of the streaming RGB-D images are automatically tracked using XMem, which predicts object masks according to both temporal and spatial correspondence of historical RGB images. 
\end{itemize}
Using TAM, there is no need to prepare datasets or train models for segmentation. Furthermore, we surprisingly find that we can re-use the annotation for an object in the annotation stage if the objects and scene backgrounds are similar, which largely reduces the effort for annotation. Although TAM can track any object with arbitrary texture, in practice we attach color markers to the gripper (See Table \ref{table:mask}) to enhance the tracking robustness in real-world scenes of RAS.

\begin{table}[!tbp]
   \scriptsize
   \renewcommand{\arraystretch}{1.3}
   \caption{Visual Masks for Surgical Grasping}
   \centering
   \begin{tabular}{c c c c}
   \Xhline{4\arrayrulewidth}
   \bf{Class ID} & \bf{Category} & \bf{Attach Marker}  & \bf{10mm Depth Accuracy}\\
   \hline
   \rowcolor{gray!20}Target & Target Object & \ding{53} & \checkmark \\
   Gripper Base & Gripper & \checkmark & \checkmark \\
   \rowcolor{gray!20}Gripper Tip & Gripper & \checkmark & \ding{53}\\
   
   \Xhline{4\arrayrulewidth}
   \end{tabular}
\label{table:mask}
   \end{table}

\color{black}

 \begin{figure*}[!tbp]
  \centering
  \includegraphics[width=1.0\hsize]{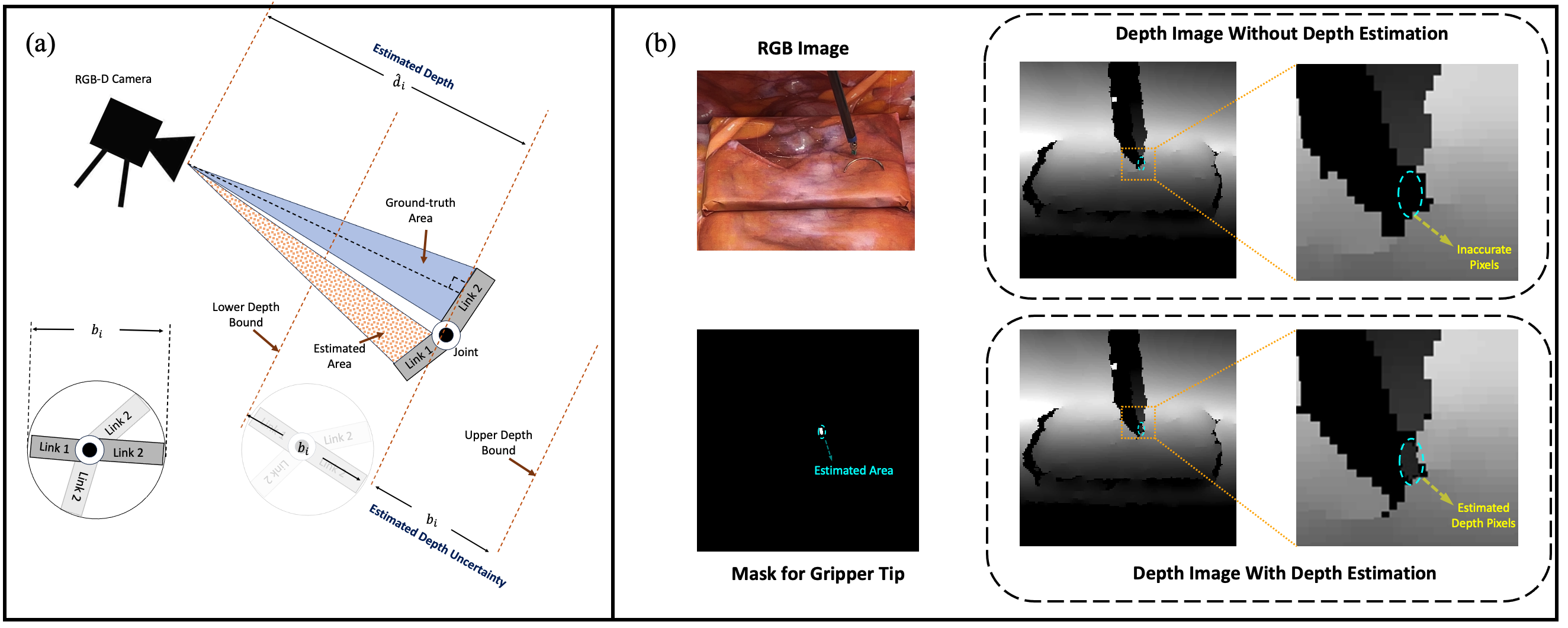}
  \caption{Schematic illustration and real-world visualization of our uncertainty-aware depth estimation. (a) We show a schematic illustration of our uncertainty-aware depth estimation for a two-link robot arm on a 2D plane. The area of significant depth noise 
  occurs due to the sensing principle of a structured light depth camera. Depth in noisy area is estimated as the depth median in the ground-truth area. The uncertainty is the minimal diameter of the spatial bound for an arbitrary object's configuration (the sum of the lengths of two links in this example). (b) We show the observed real-world RGB-D images and the mask for a gripper tip in general surgical grasping. The original depth in the gripper tip region is prone to be inaccurate (see top right). The depth pixels in the region of the gripper tip are calculated with our depth estimation (see bottom right).}
  \vspace{-0.45cm}
 \label{fig:uncertain_depth_estimation}
\end{figure*}
\color{black}
\subsection{Depth Estimation for Object's Imprecise Region} \label{sec:Depth estimation}
To facilitate our visuomotor policy, depth images of task objects with arbitrary shapes are required to be captured at each timestep. An Intel Realsense D435 RGB-D camera, a structured-light depth camera, is used to capture the RGB-D images for general surgical grasping, similar to \cite{scheikl2022sim}. However, the depth error for a small object is prone to be large since the infrared structured light pattern can not be fully projected to the object's surface due to its small and narrow size. Failing to project a complete infrared pattern leads to an imprecise depth image for the small object \cite{haider2022can}. For instance, the depth image of the gripper's tip can experience depth errors of up to 100mm in our experiments. Such discrepancy is unacceptable in surgical grasping tasks where millimeter-level accuracy is necessary.

We aim to estimate the values and uncertainties of depth pixels for a rigid-link object’s inaccurate region when the object's depth pixels are partially (but not fully) imprecise. Our intuition is that we can estimate depth pixels in the imprecise region of the object with the pixels in the closest precise region of the same object. To this end, we first empirically divide masks $M$ into two sets: estimation masks $M^{e}$ and ground-truth masks $M^{gt}$ that contain imprecise and precise depth pixels, respectively.
One can distinguish between the estimation and the ground-truth masks by the maximum pixel error in the mask region between the captured depth image from the camera and the ground-truth depth image (from eye observation or prior geometric models) within the workspace. 
In the case of our grasping tasks, the maximum pixel error is set to 10mm. 
We can easily find that only the mask of the gripper tip is estimated and the rest masks are ground truth based on the empirical depth accuracy in Table \ref{table:mask}. For each estimation mask $m_i^{e}\in M^e$, we then find the ground-truth mask $m_i^{\star}$ that has the smallest centroid distance and belongs to the same category, which is formulated as
\begin{equation}
   \begin{aligned}m_i^{\star}&=\mathop{\arg\min}_{m_j^{gt}\in M^{gt}} \ \  \mathrm{C} (m_i^{e},m_j^{gt}). \\
    &\textit{ \textit{s.t.} $m_j^{gt}$ and $m_i^{e}$ have the same categories},
   \end{aligned}
\end{equation}
where $C(\cdot,\cdot)$ is a distance function that computes the centroid pixel distance between two masks. Finally, given the closest ground-truth object mask  $m_i^{\star}$, we compute the estimated depth  $\hat{d}_i$ for the mask region $m_i^{e}$ as
\begin{equation}
    \hat{d}_i =  Med(I^{d}\odot m_i^{\star}),
\end{equation}
where $Med(\cdot)$ is a median function that computes the median of non-zero elements, $I^{d}\in \mathbb{R}^{600\times 600}$ is the depth channel of the observed RGB-D image and $\odot$ is the element-wise product operator. The estimated depth image $\hat{I}_i^d$ for mask $m^e_i$ is formulated as
\begin{equation}
    \hat{I}_i^d = U(\hat{d}_i-b_i,\hat{d}_i+b_i)\odot m^e_i,
\end{equation}
where $b_i$, denoted as empirical size, is the minimal diameter of a bounding ball that contains masks $m_i^{\star}$ and $m_i^{e}$ given an arbitrary object’s configuration, and the elements of matrix $U(\hat{d}_i-b_i,\hat{d}_i+b_i)\in \mathbb{R}^{600\times 600}$ are sampled from a uniform distribution with the lower bound $\hat{d}_i-b_i$ and the upper bound $\hat{d}_i+b_i$. The bounded range indicates the range of the estimation uncertainty. The final estimated depth image $\hat{I}^{d}$ is obtained by replacing original depth pixels with the estimated pixels of $\hat{I}_i^d$ within the mask region for all estimated masks $M^{e}$. In practice, users need to set $b_i$ as a hyper-parameter before running our depth estimation and it can be attained easily by empirical experiments. We set it as 7mm for the gripper tip mask.

\begin{figure*}[!tbp]
  \centering
  \includegraphics[width=1\hsize]{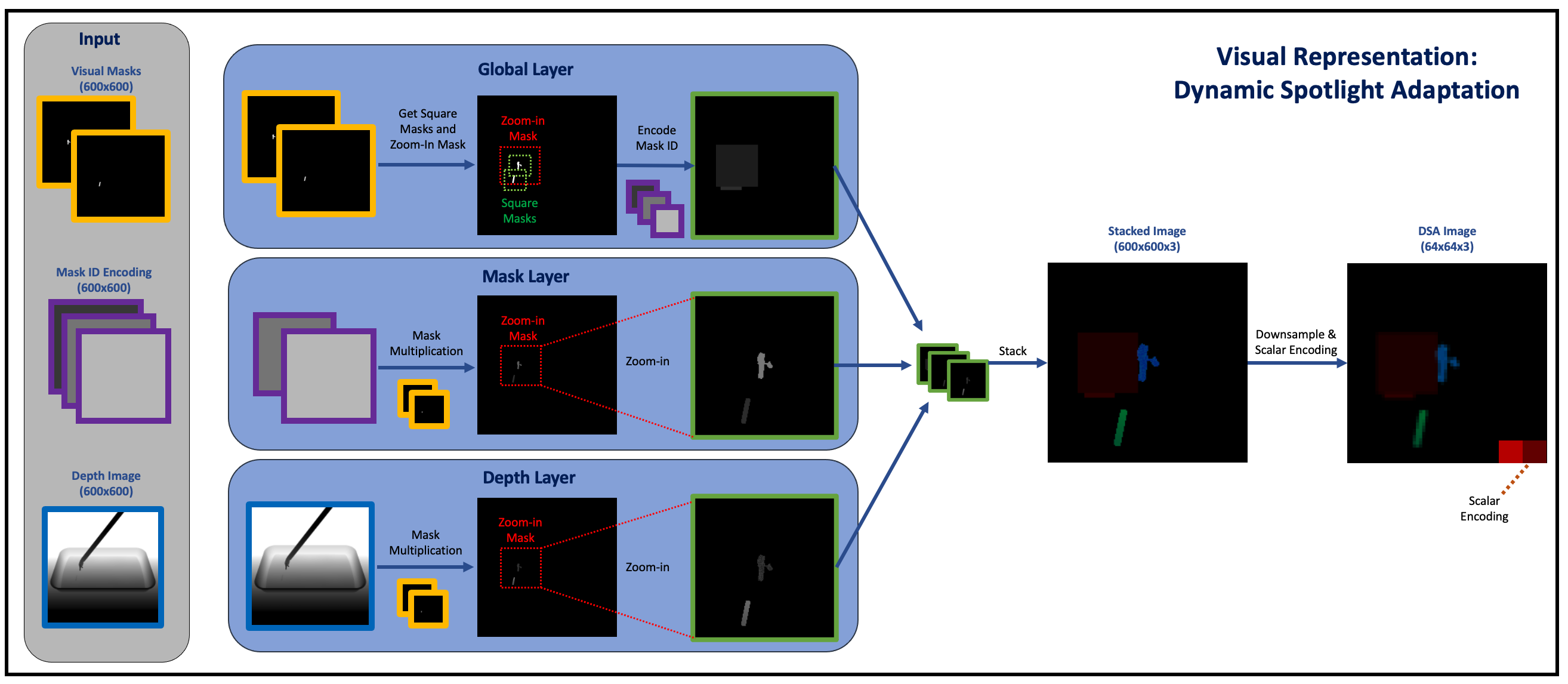}
  \caption{Pipeline of Dynamic Spotlight Adaptation (DSA) for visual representation of world models in general surgical grasping. The visual masks, encoding matrices of mask ID, and the estimated depth image are the inputs of our visual representation. The global layer (red channel) is obtained by generating square masks and a zoom-in mask based on visual masks, followed by summing these masks with encoding values. The mask (green channel) and the depth layers (blue channel) are obtained by zooming in a segmented mask image and a segmented depth image, respectively, which can be obtained by multiplication with visual masks. We stack three layers into a 3D matrix and downsample it to a 64x64x3 compact image for world model learning. Two scalar signals, i.e. the task-level state and the gripper toggling state, are encoded into two 6x6 square images, which are used to replace pixels of the downsampled image at the bottom right corner of the global layer (red channel).}
  \vspace{-0.45cm}
  \label{fig:DSA}
\end{figure*}
 
\subsection{Visual Representation} 
\label{sec:visual_representation}
Since the Variational Autoencoder (VAE) in DreamerV2 is notoriously hard to scale to high-dimensional images \cite{kingma2013auto}, a common practice is down-sampling images to a low 64x64x3 resolution \cite{hafner2020mastering}. However, such down-sampled images result in degraded visuomotor performance as shown in our experiments in Sec. \ref{sec:exp_visual_rep}.

\color{black}

Instead, we extend Dynamic Spotlight Adaptation's (DSA) compact visual representation \cite{lin2023end} to the setting of general surgical grasping. Our visual representation encodes both depth and mask images (size: 600x600) of task objects into a single compact 3-channel image (size: 64x64x3) via dynamically zooming in the mask regions, minimizing the loss of mask coordinates information and image quality of both depth and mask. Our representation only encodes visual information of two objects, the gripper tip and the target grasping object, which results in a set of DSA masks $M'=\{m_i^{'} \in\mathbb{R}^{600\times600} \}_i^{2}\subset M$, containing two corresponding masks for these objects. The visual information is extracted from global and local levels: at the global level, the object masks are represented by square masks $M_s=\{m_{s_i}\in\mathbb{R}^{600\times600}\}_{i=1}^2$, where the centroid pixel coordinates of the square mask coincide with those of the original mask and the length of the square mask is fixed at  $\alpha_{s}\in(0,1)$ ratio of the image length; we zoom in a square mask $m_{zoom}\in\mathbb{R}^{600\times600}$  to obtain local-level images, where the centroid pixel coordinates of the zoom-in mask coincide with those of the mask of the gripper tip and the length of the zoom-in mask is fixed at $\alpha_{zoom}\in(0,1)$ ratio of the image length; note that in practice the zoom-in mask is movable and automatically tracks the pixel position of the mask region of the gripper tip due to the coincidence constraint; to keep the masks from going out of sight, the position of masks are clipped to range $[1,600]$, where the mask outside the interval is translated to the nearest edge of the image to make all pixel visible; when the square mask overlaps the zoom-in mask, the elements of the overlapped region for the square mask is set to zeros; we encode the IDs of the square masks and the zoom-in mask as matrices $V=\{v_i\in\mathbb{R}^{600\times600}\}_{i=1}^2$ and $v_{zoom}\in\mathbb{R}^{600\times600}$, respectively, where the matrices are obtained by broadcasting the normalized encoded values (range: $[0, 255]$) to the matrix elements; the encoded matrices are element-wise multiplied by their corresponding masks and the resultant matrices are summed to form a global layer $I_1\in\mathbb{R}^{600\times600}$. At the local level, the encoded matrices $V$ and the estimated depth image $\hat{I}^d$ are element-wise multiplied by the corresponding original masks $M^{'}$, followed by summing their resultant matrices to form a mask-encoding matrix and a masked depth matrix, respectively; we zoom in the mask area $m_{zoom}$ of both the mask-encoding matrix and the masked depth matrix to form a local mask layer $I_2\in\mathbb{R}^{600\times600}$ and a local depth layer $I_3\in\mathbb{R}^{600\times600}$, respectively. We formulate three layers in our representation as
\begin{equation}
\begin{aligned}
    \text{Global Layer}: I_1 = &\sum_{i=1}^{|M_s|}m_{s_i}\odot v_{i}  +
     m_{zoom} \odot v_{zoom}, \\
    \text{Mask Layer}:I_2 = &f_{zoom}(\sum_{i=1}^{|M^{'}|}m_i^{'} \odot v_i, m_{zoom}),\\
    \text{Depth Layer}:I_3 =& f_{zoom}(\sum_{i=1}^{|M^{'}|}m_i^{'} \odot \hat{I}^d , m_{zoom}),
\end{aligned}
\end{equation}
where $f_{zoom}(\cdot,m_{zoom})$ is a zoom-in function that resize the local area $m_{zoom}$ of the input image to a larger 600x600 image. Our visual representation is formed by stacking these 3 layers, followed by image downsampling (image size: 64x64x3). Finally, we encode two scalar signals, the IDs of the task-level state in Sec. \ref{sec:task-level state} and the gripper toggling state: we normalize the scalar values to the range $[0, 255]$ and broadcast the normalized values to pixels of two 6x6 square images. The final DSA image can be obtained by replacing the image pixels of the global layer at the bottom right corner with the encoded pixels of the square images. The pipeline of our visual representation is shown in Fig. \ref{fig:DSA}.

 \begin{figure}[!tbp]
  \centering
  \includegraphics[width=1.0\hsize]{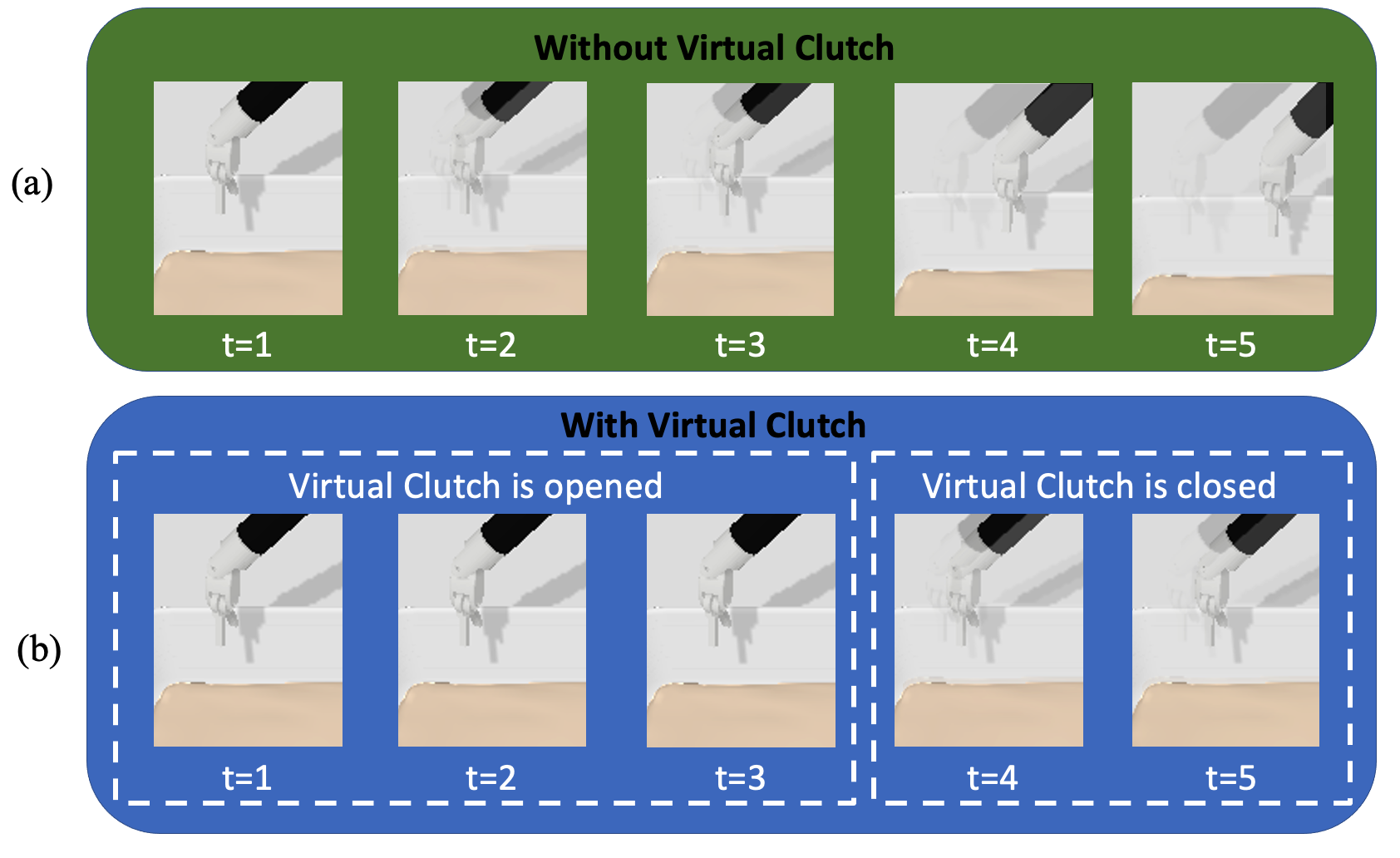}
  \caption{\color{black}Schematic illustration for virtual clutch (VC). (a) We show the first 5 timesteps of a rollout when a robot gripper is controlled to move to the right without applying VC. The transparent areas represent the starting positions of the gripper. (b) We show the corresponding 5 timesteps after applying VC ($H_{clutch}=4$). The gripper remains stationary until the fourth timestep when it begins to move. \color{black}} 
  \vspace{-0.45cm}
\label{fig:VC}
\end{figure}
\subsection{Virtual Clutch}
We deploy the technique of control processing, Virtual Clutch (VC) \cite{lin2023end}, to alleviate the issue of inaccurate estimation of posterior states. In \cite{lin2023end}, a significant error from the estimated posterior hidden state is reported at the beginning of a rollout due to poor initial prior guess (e.g., a zero matrix in \cite{hafner2019dream}) and slow convergence on the error. Such a discrepancy is mainly caused by a lack of temporal observations at the beginning. Inferring an accurate hidden state required observing a handful of visual observations starting from a poor initial prior guess. Applying VC can significantly increase the learning stability and the final performance of the visuomotor policy. 

The intuition of VC is that the agent starts to move only when it observes a sufficient amount of temporal visual observation. As such, we ensure a sufficient low estimation error when robots interact with the environment. To this end, we apply VC to learn visuomotor policy in surgical grasping (see Fig. \ref{fig:VC} for the illustration of VC). In particular, the control output of our controller $a_t^c$ is determined by a timestep-dependent ``clutch'' as
\begin{equation}
    a_t^c = \begin{cases} a_t, & t \geq H_{clutch}\\
    a_{idle}, & t<H_{clutch},
    \end{cases}
\end{equation}
where $a_{idle}$ is an idle action command that keeps the joint positions of the robot arm unchanged, $H_{clutch}$ is a non-negative constant determining the timestep that starts to close the ``clutch'', and $a_t$ is the action predicted by the learned policy.

\subsection{Domain Randomization} 

We apply domain randomization to introduce uncertainty to train visuomotor policy in simulation. In particular, we apply 4 types of noise:  
\begin{itemize}
\item \textit{Camera Pose}: The initial camera poses are randomized. A small random pose disturbance is dynamically applied to the camera pose at each timestep of a rollout, where the noise magnitude is a ratio (called dynamic noise ratio) of the feasible range magnitude of the initial camera pose. The goal is to force our learned visuomotor controller to adapt to camera pose variations and disturbances, similar to \cite{seo2023multi}.  

\item \textit{Target Disturbance}: A small pose random disturbance in the direction horizontal to the plane is applied to the pose of the target grasping object at each timestep of a rollout, where the noise magnitude is a ratio (called dynamic noise ratio) of the feasible range magnitude of the workspace. The goal is to simulate the dynamic motion of the target object in real surgery.

\item \textit{Gripper Noise}: Randomized noises are dynamically applied to all actuated directions of the gripper in Cartesian space at each timestep of a rollout, where the noise magnitude is a ratio (called dynamic noise ratio) of the feasible range magnitude of action commands. Note that the action noise can be dependent and not mutually exclusive in feasible directions of translation. The goal is to simulate the error between the measured and desired gripper poses via the end-effector control. 
\begin{table}[!tbp]
\scriptsize
\renewcommand{\arraystretch}{1.3}
\caption{Value/Range for Domain Randomization}
\centering
\begin{tabular}{c c c }
   \Xhline{4\arrayrulewidth}
   \bf{Type} & \bf{Name} & \bf{Value/Range}  \\ \hline
   \multirow{8}{*}{Camera Pose} & Roll Angle & $[-11^{o}, 11^{o}]$ \\
   & Pitch Angle & $[35^{o}, 55^{o}]$ \\
   & Yaw Angle & $[-5^{o}, 5^{o}]$ \\
   & Target Location& $[-10mm, 10mm]$ \\
   & Target Distance& $[250mm, 350mm]$ \\
   & Depth Range& $[90mm, 110mm]$ \\
   & Depth Center& $[290mm, 310mm]$ \\
   & Dynamic Noise Ratio & $[-0.1, 0.1]$ \\
   \hline
    \multirow{1}{*}{Target Disturbance} 
    & Dynamic Noise Ratio  & $[-0.01, 0.01]$ \\
    \hline
    \multirow{1}{*}{Gripper Noise} & Dynamic Noise Ratio & $[-0.04, 0.04]$ \\
    \hline
    \multirow{6}{*}{Image Noise}& SnP Amount & $0.05$ \\
     & SnP Balance& $0.5$ \\
     & Gaussian Blur Kernel Size& $3$ \\
     & Gaussian Blur Sigma& $0.8$ \\
     & RGB-D Cutout Amount& $[0.0,0.46]$ \\

   \hline
\Xhline{4\arrayrulewidth}
\end{tabular}
\label{table:domain_random}
\end{table}
\item \textit{Image Noise}: Simulated random salt and pepper (SnP) noise, Gaussian blur, and cutouts are added to the observed RGB-D images in simulation, as  \cite{horvath2022object} suggest. Box-shape and circle-shape cutouts with random sizes, positions, and colors, are used to simulate occlusions. We control the noise magnitude of SnP and cutouts by amount coefficients, indicating the probability of occurrence. In addition, the occurrence ratio of salt and pepper noises is defined as a balance coefficient. The Gaussian blur is controlled by the size of the Gaussian kernel and the variance, sigma. 
\end{itemize}
\color{black}Detailed parameters of our domain randomization are shown in Table \ref{table:domain_random}.\color{black}

\subsection{FSM-Driven Reward and Termination}
\label{sec:task-level state}
Constructing a shaped reward function $R(s, a)$, known as \textit{dense reward}, is challenging for general surgical grasping since it necessitates expert-level domain knowledge and cumbersome engineering  \cite{lin2023end}. Furthermore, dense rewards in previous works of surgical grasping require tracking either object poses or feature coordinates, which is not available in our assumption. 

We use discrete finite \textit{sparse rewards} to alleviate such limitations. In particular, four \textit{task-level states}, directly determining both rewards and terminations, are defined as
\begin{itemize}
    \item \textit{Successful Termination}: When the target object is successfully grasped and lifted, a reward of $1$ is given and the task is terminated.
    \item \textit{Failed Termination}: When the current timestep reaches the horizon limit, a reward of $-0.1$ is given and the task is terminated.
    \item \textit{Normal Progress}: When the task proceeds normally without termination flags, a small reward of $-0.001$ is given to speed up the grasping. The task is not terminated in such a task-level state.
    \item \textit{Abnormal Progress}: When any of the following three abnormalities occur: i) the gripper's desired pose exceeds the workspace; ii) the target object slides horizontally due to external forces from the gripper; or iii) the target object is not in the zoom-in box of DSA, negative rewards are provided. Particularly, they are set to $-0.01$, $-0.01$, and $-0.005$ to discourage the occurrence of these events, respectively. In this task-level state, the task is not terminated.
\end{itemize}
Since these task-level states are finite and their transitions are fully defined, we can use a finite-state machine (FSM) to monitor these states. Each task-level state is stored by an FSM's state. The transition of the FSM's states is based on that of their stored task-level state. Note that the agent can not determine the termination of the task due to our FSM-driven mechanism.

\begin{figure}[!tbp]
  \centering
  \includegraphics[width=\hsize]{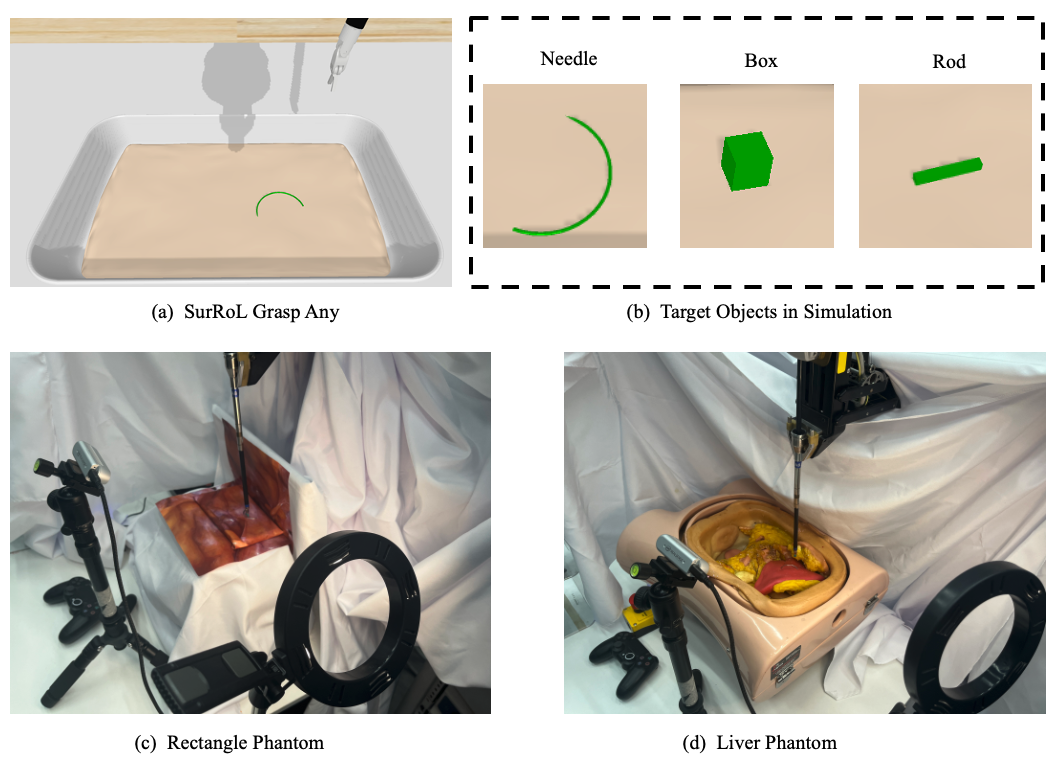}
  \caption{Experiement setups and simulated target objects. We show the simulation setup in (a) and two real-robot setups, including a setup with rectangle phantom in (c) and a setup with liver phantom in (d). Simulated target objects are shown in (b).}
  \vspace{-0.45cm}
\label{fig:exp_setup}
\end{figure}
\section{Experiment Setup}
\subsection{Environment}
\color{black}The experiments of surgical grasping tasks are conducted on both a simulation and a real robot. We develop a simulation benchmark (See Fig. \ref{fig:exp_setup}a) and two real-robot benchmarks (See Fig. \ref{fig:exp_setup}c-d). 

\color{black}For simulation, the benchmark is developed based on an open-source simulation platform for surgical robot learning, SurRoL\cite{xu2021surrol}. 3 types of grasping objects (See Fig. \ref{fig:exp_setup}b), including needles, boxes, and rods, occur in the task with equal probabilities. We also scale the shape of grasping objects from 0.75 to  1.5 randomly.

For the real robot, we deploy learned controllers to a surgical robotic arm, Patient Side Manipulator (PSM), in da Vinci Surgical System controlled by an open-source control system, da Vinci Research Kit (dVRK) \cite{kazanzides2014open}. Observed RGB-D images are captured by an Intel RealSense D435 camera. A filling light is used to control the illumination condition. The transformation from the camera frame to the gripper frame, known as the hand-eye transformation, is set within the range of our domain randomization. A flag signal, indicating successful grasping, is manually sent by a joystick based on human observation. Our FSM transit to the FSM's state of successful termination when the grasping flag is sent and the height of the gripper's desired pose is above 10mm. We evaluate controllers on two phantoms, a rectangle phantom, and a liver phantom (See Fig. \ref{fig:exp_setup}c-d). The rectangle phantom, which is easier to control experimental variables, is used for the studies of performance, generality, and robustness, while the liver phantom, providing more realistic shapes and mechanical properties, is used in our performance study\color{black}.

\color{black}

  \begin{table*}
    \centering
    \caption{Evaluating 40 Rollouts in Simulation Over 3 Training Seeds}
        \renewcommand{\arraystretch}{1.3}
        \fontsize{7pt}{7pt}\selectfont
    \begin{tabular}{cccccccccccccccc}

   \Xhline{4\arrayrulewidth}
     \multicolumn{2}{c}{} & \multicolumn{2}{c}{GAS(Ours)} & \multicolumn{2}{c}{GAS-RawVR} & \multicolumn{2}{c}{GAS-NoDE} & \multicolumn{2}{c}{GAS-NoClutch} & \multicolumn{2}{c}{GAS-NoDR} & \multicolumn{2}{c}{PPO}  \cite{schulman2017proximal}
     & \multicolumn{2}{c}{DreamerV2} \cite{hafner2019dream} \\
     \rowcolor{gray!20} & Target Object  & SR & Score & SR & Score & SR & Score & SR & Score & SR & Score & SR & Score & SR & Score  \\ \hline
      
    & Needle & \textbf{86} $\pm$ \textbf{4} & \textbf{52} $\pm$ \textbf{24} & $0\pm 0$ & $0\pm 0$&$0\pm 0$&$0\pm 0$ & $32\pm 32$ & 17 $\pm$ 28 & 78 $\pm$ 3 & 47 $\pm$ 31 & 0 $\pm$ 0 & 0 $\pm$ 0 & 0 $\pm$ 0 & 0 $\pm$ 0\\

    & box\&rod & \textbf{89} $\pm$ \textbf{7} & \textbf{60} $\pm$ \textbf{18}&$0\pm 0$&$0\pm 0$& $0\pm 0$& $0\pm 0$ & 40 $\pm$ 40 & 26 $\pm$ 32 &  $85 \pm 4$ & 51 $\pm$ 24 &0 $\pm$ 0 & 0 $\pm$ 0 & 0 $\pm$ 0 & 0 $\pm$ 0\\

    & Aggregate & \textbf{87} $\pm$ \textbf{6} & \textbf{56} $\pm$ \textbf{21} & $0\pm 0$ & $0\pm 0$ & $0\pm 0$ & $0\pm 0$ &  36 $\pm$ 36& 22 $\pm$ 31 & 84 $\pm$ 6 &49 $\pm$ 28 &0 $\pm$ 0 & 0 $\pm$ 0& 0 $\pm$ 0 & 0 $\pm$ 0\\
     \Xhline{4\arrayrulewidth}
        \label{table:simulation}
\end{tabular}
\end{table*}
\subsection{Baselines and Ablations}
SOTA pixel-level RL methods, \texttt{PPO} \cite{schulman2017proximal} and \texttt{DreamerV2} \cite{hafner2019dream}, are compared in our experiments. In these baselines, visual signals (i.e., RGB-D observed images, object masks) and scalar signals (i.e., gripper states, and task-level states) are directly fed as observation. \color{black} A visual representation, raw visual representation (RawVR), is used in these baselines to encode RGB-D images and object masks: we stack 3 images, including i) the depth image, ii) the encoded mask image, and iii) a grayscale image converted from the RGB image, to form a 3-channel image; the stacked image is downsampled to $64\times 64\times3$ resolution for a fair comparison with our DSA.
\color{black} For the PPO baseline, we follow the implementation in \cite{wu2023daydreamer}: the scalar signals are broadcasted to image planes and concatenate them with RGB-D channels, a common practice in \cite{schrittwieser2020mastering}; the resultant image is encoded to a compact vector via a pre-trained universal visual encoder, R3M \cite{nair2022r3m}; a multilayer perception (MLP) policy leverages the vectorized encoding as input and is learned by the standard PPO framework \cite{schulman2017proximal}. For DreamerV2, a convolutional neural network and an MLP are used to encode the RGB-D images and the scalar signals, respectively, which is a standard implementation in \cite{hafner2020mastering}.\color{black} 

We also compare with ablation baselines to evaluate the effectiveness of our proposed modules:
\begin{itemize}
   \item \texttt{GAS-RawVR}: 
   \color{black}
   We replace our visual representation with RawVR. Scalar signals are encoded to 6x6 square images, replacing the image pixels in the $1^{st}$ channel, similar to our visual representation. \color{black}
   \item \texttt{GAS-NoDE}: We use the original depth image from the observed RGB-D image instead of the depth image inferred by our depth estimation.
   \item \texttt{GAS-NoClutch}: We remove the effect of VC to evaluate its effectiveness, which can be achieved by setting $H_{clutch}$ to $1$.
   \item \texttt{GAS-NoDR}: We remove the effect of domain randomization during training to evaluate its effectiveness in sim-to-real transfer. 
 
\end{itemize}

\subsection{Training and Evaluation}

For evaluation, we compare two metrics: i) the success rate as a traditional indicator in surgical grasping  \cite{xu2021surrol,chiu2021bimanual,wilcox2022learning} and ii) \color{black} a novel indicator, \textit{Grasping Score} $s_{g}$, to evaluate how fast the controller achieves successful grasping. We define the score as 
\begin{equation}
\label{equ:score}
    s_{g} = (T - H) / H,
\end{equation}
where $T$ is the terminated timestep and $s_g\in[0,1]$. \color{black} 

Controllers are trained in the simulation benchmark. In the training process, controllers are trained with $1.8$ million timesteps; $50$ evaluation rollouts are carried out every $200K$ training timestep; we pre-fill $10K$ timesteps of rollouts before training, including $3K$ timesteps using a scripted demonstration policy (approximately 20 rollouts) and $7K$ timesteps with a random policy. Each controller is trained with 3 seeds. We train on 5 computers with 8G-RAM or above GPUs, where the minimum computer configuration is a laptop with a 12G-RAM GPU, RTX 4070, and an 8-core CPU. Each simulation experiment consumes 2-3 days for training. We select the controller with the highest evaluation score among the 3 seeds to deploy on the real robot. In the real robot, the evaluated controllers run at around 2 Hz, where the computer configuration is a desktop with a 24G-RAM GPU, Nvidia TITAN RTX, and an 8-core CPU. Detailed hyper-parameters for our method can be found in Table \ref{table:hyperparam} of our appendix section.

 \begin{figure}[!tbp]
  \centering
  \includegraphics[width=\hsize]{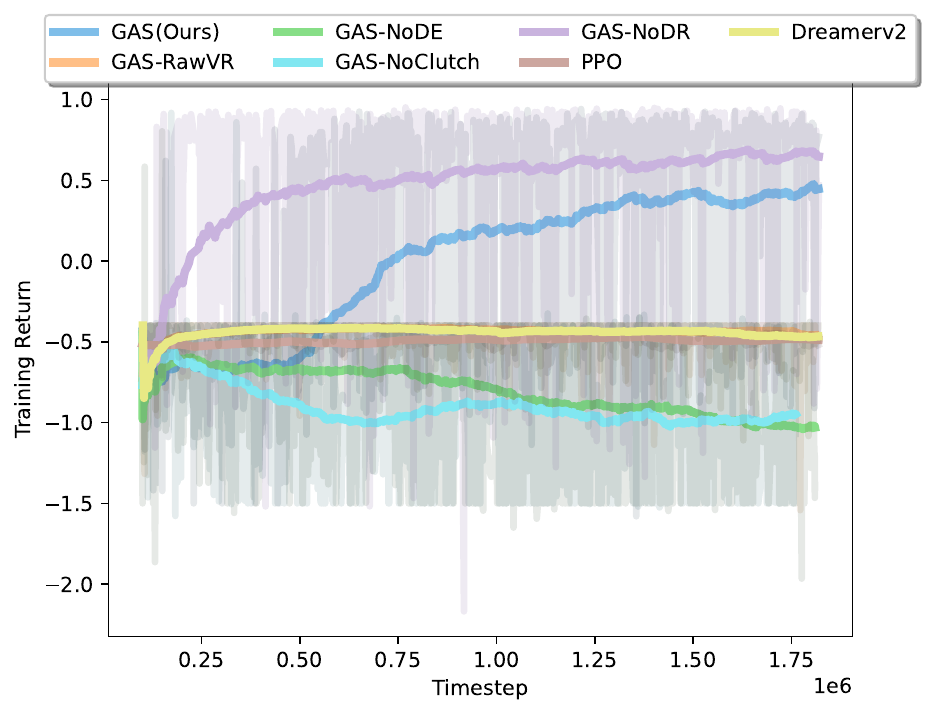}
  \caption{Training Return in Simulation. Curves are smoothed by Exponential Moving Window with a $0.99$ smoothing factor.}
  \vspace{-0.45cm}
  \label{fig:training_curve}
\end{figure}

\section{Results}
\subsection{Simulation}
 \subsubsection{Depth Estimation}
We first investigate the effectiveness of our uncertainty-aware object depth estimation. In Fig. \ref{fig:training_curve}, we observe that the baseline without depth estimation (indicated by the green curve) degrades at the beginning of the training timesteps. The performance continues to decline as the training step increases. The main reason for the performance degradation is the lack of depth accuracy for task objects (especially for the gripper tip) due to the limitation of Intel Realsense D435. After applying our depth estimation, the learning curve (in blue) starts to incline after $500K$ timesteps. \color{black}The results show that our depth estimation can boost the learned controller's performance by increasing depth accuracy.\color{black}
\subsubsection{Visual Representation}
\label{sec:exp_visual_rep}
Next, we investigate the effectiveness of two visual representations: RawVR and our DSA. From the observation of Fig. \ref{fig:training_curve}, we find that the controller with RawVR representation (see the orange curve) does not improve during training. The phenomenon indicates that the RawVR, leveraging the original gray image, depth image, and visual masks, is extremely challenging to learn useful visual features via auto-encoding of world models in surgical grasping. The reasons for such a difficulty might be i) the encoders struggle to learn small-region task-related visual features due to small gradients of image loss compared to that of the background, ii) the image quality degrades dramatically after applying image down-sampling. By explicitly removing the background image and zooming in the task-related region dynamically, DSA drastically reduces the learning difficulty of the encoder and thus improves the controller's performance. 

\subsubsection{Virtual Clutch}
\label{sec:Virtual_Clutch_exp}
Then we investigate the effectiveness of VC. In Fig. \ref{fig:training_curve}, we observe that the training curve in cyan, indicating GAS-NoClutch, is the earliest to degrade among all controllers' curves. The performance degradation is consistent with the experiment result in needle picking \cite{lin2023end}. This phenomenon indicates the training instability using world models. We conjecture that it is caused by the large error of the estimated posterior state at the beginning timesteps of a rollout due to poor initial prior guess. After applying the VC, the training stability is significantly improved (see the blue curve for the case with the VC). We will study the reasons for the training instability systematically in the future.
 \begin{figure}[!tbp]
 
  \centering
  \includegraphics[width=\hsize]{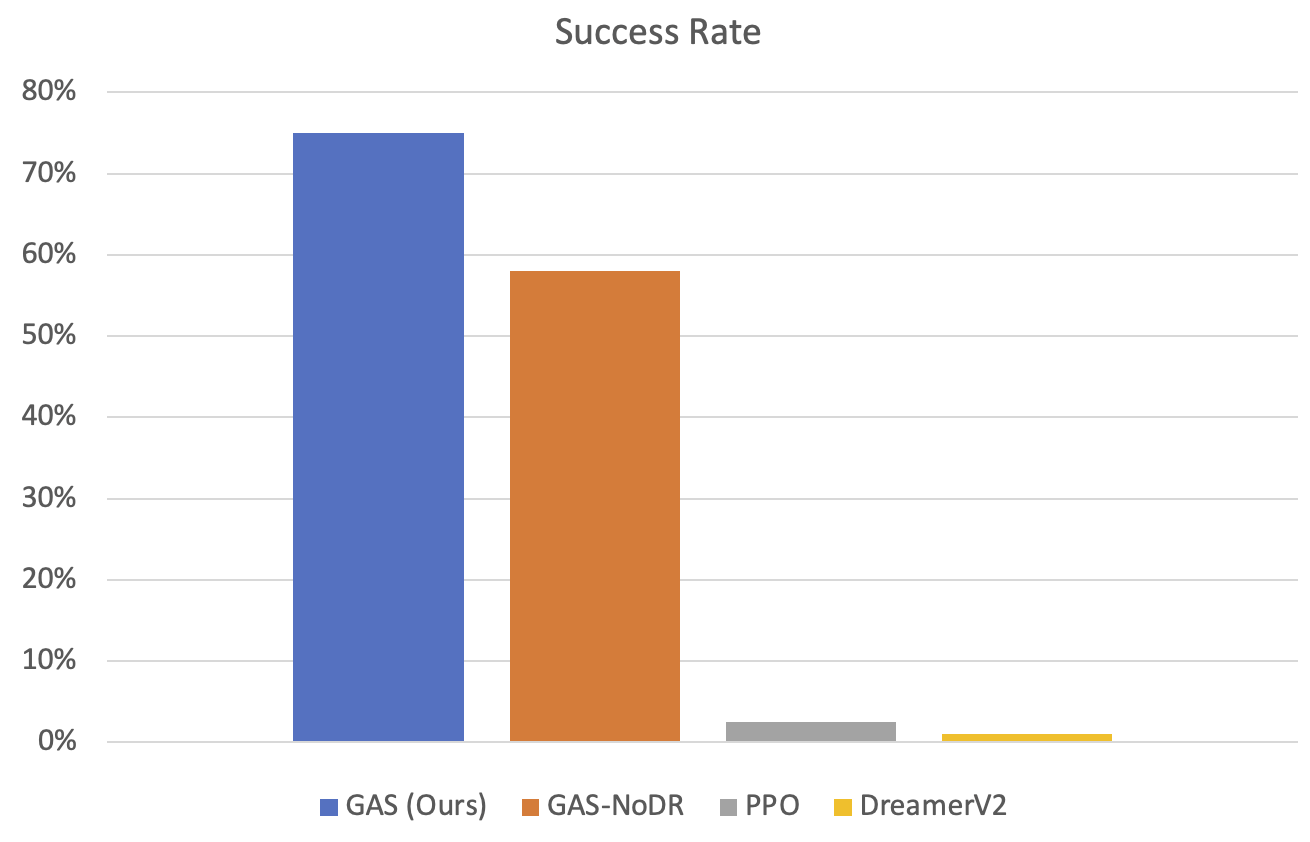}
  \caption{Performance study on rectangle phantom in a real robot.}
  \vspace{-0.45cm}
  \label{fig:dvrk_performance}
\end{figure}

\color{black}
\subsubsection{Comparisons With SOTA Methods}

Finally, two SOTA methods, DreamerV2 and PPO, are compared with our method. We observe that both DreamerV2 and PPO have not improved due to the difficulty in learning surgical grasping based on Fig. \ref{fig:training_curve}. Either DreamerV2 or PPO does not learn to perform a successful grasp in the evaluation of simulation (See Table \ref{table:simulation}). Compared with SOTA methods, our method is more sampling efficient and achieves much better performance given limited training data, which achieves $87\%$ success rate (SR) and a score of $56$ in simulation given $1.8$ million training timesteps. Yet we believe the performance of the baseline SOTA methods will start to improve if more training steps are given. The convergence performance of the baseline methods will be studied in the future.
\begin{table}[!tbp]
\color{black}
   \scriptsize
   \renewcommand{\arraystretch}{1.3}
   \caption{Studies of Performance, General and Robustness for GAS in Real Robot}
   \centering
   \begin{tabular}{c c c c c c}
   \Xhline{4\arrayrulewidth}  
   
   &\bf{Type} & \bf{Description} & \bf{Episode} & \bf{SR} (\%)  & \bf{Score} (\%) \\
    \Xhline{2\arrayrulewidth}
      \multirow{2}{*}{Performance} 
   & Phantom & Rectangle & 40& 75& 53 \\
\cline{2-6}
 &  Phantom & Liver & 40& 70& 45 \\
   \Xhline{2\arrayrulewidth}
   \multirow{8}{*}{Generality} 
   &Needle &20mm Needle & 20& 75& 53 \\
\cline{2-6}
 & \multirow{3}{*}{Thread} & Long Thread & 10 & 70 & 37\\
   & & Short Thread & 10 & 70 & 37  \\
    \cline{3-6}
   &  &Aggregate & 20& 70& 37 \\
    \cline{2-6}
   &  \multirow{1}{*}{Gauze} & - & 20 & 80& 54\\
     
    \cline{2-6}
    & Sponge & - & 20 &60 & 33\\
   \cline{2-6}
    & Fragment & Raisin & 20 & 60&39\\
\cline{2-6}
    & Gripper & - & 20 & 70&39\\
\cline{2-6}
& Aggregate & - & 120 & 69 & 43\\
\Xhline{2\arrayrulewidth}
    \multirow{10}{*}{Robustness} &\multirow{3}{*}{Background} & Background 2  & 10 & 70& 37\\
   & & Background 3  & 10 & 80& 32\\
    \cline{3-6}
   & &Aggregate & 20& 75& 35 \\
\cline{2-6}
 & \multirow{3}{*}{Target Noise} &Intermittence & 10 & 70 & 44\\
   & & Back\&Forth  & 10 & 90 & 48  \\
   \cline{3-6}
   &  &Aggregate & 20& 80& 46 \\
    \cline{2-6}
   &  Camera Pose & Random Pose & 20 & 75& 37\\
     
    \cline{2-6}
    & Action Noise & - & 20 & 75& 43\\
   \cline{2-6}
    & Image Noise & - & 20 & 75&35\\
   \cline{2-6}
      & Re-grasping & - & 20 & 55& 12\\
   
   \Xhline{4\arrayrulewidth}
   \end{tabular}
   \label{table:dvrk_studies}
\color{black}
   \end{table}
\subsection{Real Robot}
\subsubsection{Performance}
\begin{figure}[!tbp]
  \centering
  \includegraphics[width=\hsize]{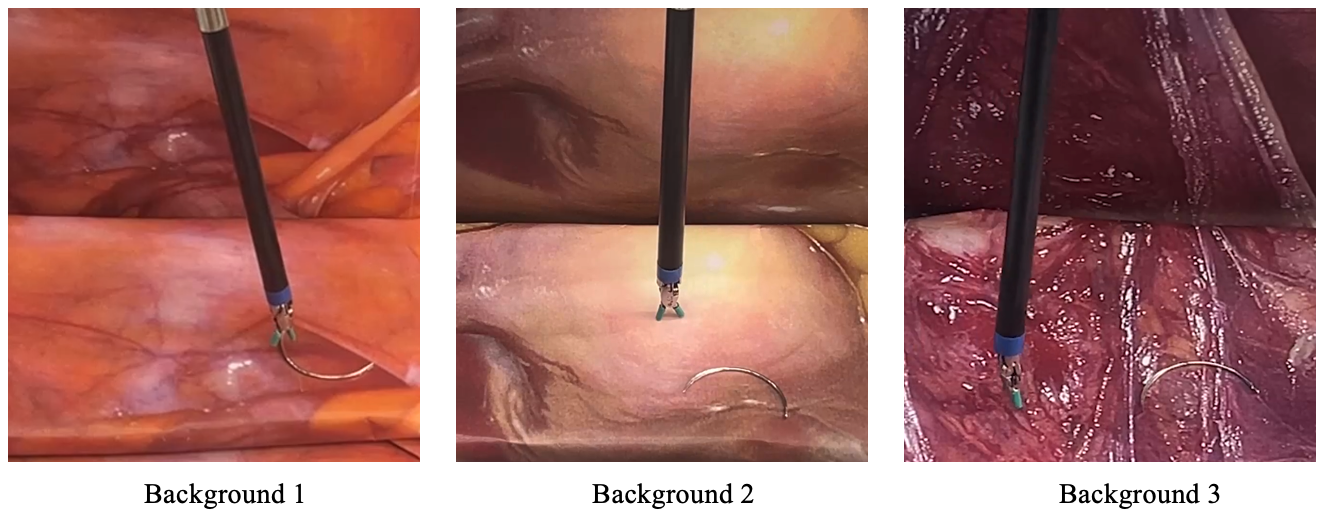}
  \caption{Backgrounds in our generality study.}
  \vspace{-0.45cm}
  \label{fig:background}
\end{figure}

\color{black}
First, we compare the performance of SOTA methods, GAS-NoDR, and our method in the real robot. 40 rollouts are evaluated for each method. A standard 40mm needle is chosen as the target grasping object and Background 1 of the rectangle phantom in Fig. \ref{fig:background} is selected for evaluation.
Fig. \ref{fig:dvrk_performance} shows the success rate on the rectangle phantom. We observe that our methods, GAS, and GAS-NoDR, significantly outperform the PPO and DreamerV2. \color{black}In particular, our methods with and without domain randomization achieve success rates of $75\%$ and $58\%$, respectively, while the success rates of both PPO and DreamerV2 are below $1\%$. We observe that the learned behaviors of PPO and DreamerV2 are similar to those of random policies, indicating that these methods are still in the exploration phase due to sampling inefficiency in training. In Table \ref{table:dvrk_studies}, we show a similar performance of our controller on the liver phantom (success rate: $70\%$) compared to that on the rectangle phantom.
\subsubsection{Generality}
Next, we investigate the generality of our method in the real robot. We apply our controller to grasp diverse classes of objects with different grippers. To quantify the generality, we evaluate 5 unseen target objects and an unseen gripper as shown in Fig. \ref{fig:surgical_objects}. 20 episodes are evaluated for each unseen object. For the unseen gripper, we choose the standard 40mm needle as the target grasping object. Results are shown in Table \ref{table:dvrk_studies} and Fig. \ref{fig1}a. We observe that our controller learns to grasp diverse surgical objects with over $60\%$ success rate. Among these objects, grasping the sponge and fragments is the hardest. This is mainly caused by two factors: i) they require a higher demand for vision and control accuracy due to their small size, and ii) their weights are too small such that they slide easily in the horizontal direction due to the external force from the gripper. Our controller can also grasp non-surgical objects, as shown in our supplementary video. \color{black}

\subsubsection{Robustness}
\color{black}
Finally, we investigate our controller's robustness in the real robot quantitatively and qualitatively. The standard 40mm needle is chosen as the target grasping object. In the quantitative study, 6 classes of system disturbances are evaluated: background variation, image noise, action noise, target disturbance, camera pose variation, and re-grasping the target objects dropped from the gripper. In particular, we evaluated our controller with two different RAS backgrounds, Background 2 and 3 shown in Fig. \ref{fig:background}; the camera is reset to a new pose manually for each evaluated rollout to vary the camera pose; two types of synthetic noises, image noise and action noise, are generated by our domain randomization and added to the observed RGB-D images and desired gripper pose, respectively; for target disturbance, we use an automatic rotating platform to move the ground plane with intermittent and back-and-forth movements so to simulate the target movements placing on the patient's tissues in real surgery; in the re-grasping experiments, we manually hit the grasped needle when the first successful grasp occurs, the target object was dropped to the plane with a random pose due to the hitting and the task is successful when the gripper grasp and lift the target object for the second time within the time limit.

In the results of Table \ref{table:dvrk_studies} and Fig. \ref{fig1}b, our controller re-grasp the dropped target with a significant success rate $55\%$; we also observe that our controller does not suffer from the decrease in success rate caused by the other 5 types of disturbance, maintaining success rates above 75\%.  We also conduct extensive qualitative studies in other types of disturbances, including hand-shaking camera, occlusion of hand, blood as well as tissue, variation in the light condition, and simulated lens fogging. Details can be found in our supplementary video.
\color{black}

\section{Conclusion, Limitation, and Discussion}
In this paper, we propose a deep RL framework, GAS, to learn a pixel-level visuomotor policy that grasps diverse unseen surgical objects with different grippers and handles disturbances for surgical robots.
In particular, we propose to segment masks for general surgical grasping using TAM; a novel method is proposed to estimate the values and uncertainties of depth pixels for a rigid-link object’s inaccurate region based on the empirical prior of the object’s size; we also propose a novel visual representation, DSA, for the world model learning in general surgical grasping; VC and domain randomization are applied to increase the learning stability and performance of sim-to-real transfer of our learned controller, respectively; FSM is applied to monitor the transition of task-level states, providing sparse rewards and terminations. Extensive experiments in both simulation and the real robot are conducted to evaluate controllers' performance, generality, and robustness.

Yet, we consider the limitations of our method as follows: i) failure cases occur due to the reason that the target object is out of the workspace after the gripper's horizontal pushing;
ii) the movement speed of the control gripper is limited by a small incremental action step when using the discrete action space; iii) the identification of the FSM's states is based on human observation and control.

In our future work, we will explore the following directions to improve GAS: i) increasing the robustness of VOS, ii) deploying continuous action space for world-model-based controllers to speed up the motion, iii) reducing the occurrence of target sliding, iv) extending GAS to surgical tasks, such as thread looping, and v) automating the FSM's identification.

\bibliographystyle{IEEEtran}
\bibliography{main}

\begin{thebibliography}{10}
\providecommand{\url}[1]{#1}
\csname url@samestyle\endcsname
\providecommand{\newblock}{\relax}
\providecommand{\bibinfo}[2]{#2}
\providecommand{\BIBentrySTDinterwordspacing}{\spaceskip=0pt\relax}
\providecommand{\BIBentryALTinterwordstretchfactor}{4}
\providecommand{\BIBentryALTinterwordspacing}{\spaceskip=\fontdimen2\font plus
\BIBentryALTinterwordstretchfactor\fontdimen3\font minus \fontdimen4\font\relax}
\providecommand{\BIBforeignlanguage}[2]{{%
\expandafter\ifx\csname l@#1\endcsname\relax
\typeout{** WARNING: IEEEtran.bst: No hyphenation pattern has been}%
\typeout{** loaded for the language `#1'. Using the pattern for}%
\typeout{** the default language instead.}%
\else
\language=\csname l@#1\endcsname
\fi
#2}}
\providecommand{\BIBdecl}{\relax}
\BIBdecl

\bibitem{d2018automated}
C.~D'Ettorre \emph{et~al.}, ``Automated pick-up of suturing needles for robotic surgical assistance,'' in \emph{2018 IEEE International Conference on Robotics and Automation (ICRA)}.\hskip 1em plus 0.5em minus 0.4em\relax IEEE, 2018, pp. 1370--1377.

\bibitem{lu2020dual}
S.~Lu, T.~Shkurti, and M.~C. {\c{C}}avu{\c{s}}o{\u{g}}lu, ``Dual-arm needle manipulation with the da vinci{\textregistered} surgical robot,'' in \emph{2020 International Symposium on Medical Robotics (ISMR)}.\hskip 1em plus 0.5em minus 0.4em\relax IEEE, 2020, pp. 43--49.

\bibitem{ozguner2021visually}
O.~{\"O}zg{\"u}ner, T.~Shkurti, S.~Lu, W.~Newman, and M.~C. {\c{C}}avu{\c{s}}o{\u{g}}lu, ``Visually guided needle driving and pull for autonomous suturing,'' in \emph{2021 IEEE 17th International Conference on Automation Science and Engineering (CASE)}.\hskip 1em plus 0.5em minus 0.4em\relax IEEE, 2021, pp. 242--248.

\bibitem{schwaner2021autonomous}
K.~L. Schwaner, I.~Iturrate, J.~K. Andersen, P.~T. Jensen, and T.~R. Savarimuthu, ``Autonomous bi-manual surgical suturing based on skills learned from demonstration,'' in \emph{2021 IEEE/RSJ International Conference on Intelligent Robots and Systems (IROS)}.\hskip 1em plus 0.5em minus 0.4em\relax IEEE, 2021, pp. 4017--4024.

\bibitem{xu2021surrol}
J.~Xu \emph{et~al.}, ``Surrol: An open-source reinforcement learning centered and dvrk compatible platform for surgical robot learning,'' in \emph{2021 IEEE/RSJ International Conference on Intelligent Robots and Systems (IROS)}.\hskip 1em plus 0.5em minus 0.4em\relax IEEE, 2021, pp. 1821--1828.

\bibitem{chiu2021bimanual}
Z.-Y. Chiu \emph{et~al.}, ``Bimanual regrasping for suture needles using reinforcement learning for rapid motion planning,'' in \emph{2021 IEEE International Conference on Robotics and Automation (ICRA)}.\hskip 1em plus 0.5em minus 0.4em\relax IEEE, 2021, pp. 7737--7743.

\bibitem{bendikas2023learning}
R.~Bendikas, V.~Modugno, D.~Kanoulas, F.~Vasconcelos, and D.~Stoyanov, ``Learning needle pick-and-place without expert demonstrations,'' \emph{IEEE Robotics and Automation Letters}, 2023.

\bibitem{long2023human}
Y.~Long, W.~Wei, T.~Huang, Y.~Wang, and Q.~Dou, ``Human-in-the-loop embodied intelligence with interactive simulation environment for surgical robot learning,'' \emph{IEEE Robotics and Automation Letters}, 2023.

\bibitem{huang2023demonstration}
T.~Huang, K.~Chen, B.~Li, Y.-H. Liu, and Q.~Dou, ``Demonstration-guided reinforcement learning with efficient exploration for task automation of surgical robot,'' \emph{arXiv preprint arXiv:2302.09772}, 2023.

\bibitem{sen2016automating}
S.~Sen, A.~Garg, D.~V. Gealy, S.~McKinley, Y.~Jen, and K.~Goldberg, ``Automating multi-throw multilateral surgical suturing with a mechanical needle guide and sequential convex optimization,'' in \emph{2016 IEEE international conference on robotics and automation (ICRA)}.\hskip 1em plus 0.5em minus 0.4em\relax IEEE, 2016, pp. 4178--4185.

\bibitem{sundaresan2019automated}
P.~Sundaresan \emph{et~al.}, ``Automated extraction of surgical needles from tissue phantoms,'' in \emph{2019 IEEE 15th International Conference on Automation Science and Engineering (CASE)}.\hskip 1em plus 0.5em minus 0.4em\relax IEEE, 2019, pp. 170--177.

\bibitem{wilcox2022learning}
A.~Wilcox \emph{et~al.}, ``Learning to localize, grasp, and hand over unmodified surgical needles,'' in \emph{2022 International Conference on Robotics and Automation (ICRA)}.\hskip 1em plus 0.5em minus 0.4em\relax IEEE, 2022, pp. 9637--9643.

\bibitem{joglekar2023suture}
N.~Joglekar, F.~Liu, R.~Orosco, and M.~Yip, ``Suture thread spline reconstruction from endoscopic images for robotic surgery with reliability-driven keypoint detection,'' in \emph{2023 IEEE International Conference on Robotics and Automation (ICRA)}.\hskip 1em plus 0.5em minus 0.4em\relax IEEE, 2023, pp. 4747--4753.

\bibitem{lu2019surgical}
B.~Lu, H.~K. Chu, K.~Huang, and J.~Lai, ``Surgical suture thread detection and 3-d reconstruction using a model-free approach in a calibrated stereo visual system,'' \emph{IEEE/ASME Transactions on Mechatronics}, vol.~25, no.~2, pp. 792--803, 2019.

\bibitem{kehoe2014autonomous}
B.~Kehoe, G.~Kahn, J.~Mahler, J.~Kim, A.~Lee, A.~Lee, K.~Nakagawa, S.~Patil, W.~D. Boyd, P.~Abbeel \emph{et~al.}, ``Autonomous multilateral debridement with the raven surgical robot,'' in \emph{2014 IEEE International Conference on Robotics and Automation (ICRA)}.\hskip 1em plus 0.5em minus 0.4em\relax IEEE, 2014, pp. 1432--1439.

\bibitem{seita2018fast}
D.~Seita, S.~Krishnan, R.~Fox, S.~McKinley, J.~Canny, and K.~Goldberg, ``Fast and reliable autonomous surgical debridement with cable-driven robots using a two-phase calibration procedure,'' in \emph{2018 IEEE International Conference on Robotics and Automation (ICRA)}.\hskip 1em plus 0.5em minus 0.4em\relax IEEE, 2018, pp. 6651--6658.

\bibitem{fan2024learn}
K.~Fan, Z.~Chen, G.~Ferrigno, and E.~De~Momi, ``Learn from safe experience: Safe reinforcement learning for task automation of surgical robot,'' \emph{IEEE Transactions on Artificial Intelligence}, 2024.

\bibitem{hwang2022automating}
M.~Hwang, J.~Ichnowski, B.~Thananjeyan, D.~Seita, S.~Paradis, D.~Fer, T.~Low, and K.~Goldberg, ``Automating surgical peg transfer: calibration with deep learning can exceed speed, accuracy, and consistency of humans,'' \emph{IEEE Transactions on Automation Science and Engineering}, vol.~20, no.~2, pp. 909--922, 2022.

\bibitem{zhong2019dual}
F.~Zhong \emph{et~al.}, ``Dual-arm robotic needle insertion with active tissue deformation for autonomous suturing,'' \emph{IEEE Robotics and Automation Letters}, vol.~4, no.~3, pp. 2669--2676, 2019.

\bibitem{kalashnikov2018qt}
D.~Kalashnikov, A.~Irpan, P.~Pastor, J.~Ibarz, A.~Herzog, E.~Jang, D.~Quillen, E.~Holly, M.~Kalakrishnan, V.~Vanhoucke \emph{et~al.}, ``Qt-opt: Scalable deep reinforcement learning for vision-based robotic manipulation,'' \emph{arXiv preprint arXiv:1806.10293}, 2018.

\bibitem{seo2023multi}
Y.~Seo, J.~Kim, S.~James, K.~Lee, J.~Shin, and P.~Abbeel, ``Multi-view masked world models for visual robotic manipulation,'' \emph{arXiv preprint arXiv:2302.02408}, 2023.

\bibitem{ha2018world}
D.~Ha and J.~Schmidhuber, ``World models,'' \emph{arXiv preprint arXiv:1803.10122}, 2018.

\bibitem{wu2023daydreamer}
P.~Wu, A.~Escontrela, D.~Hafner, P.~Abbeel, and K.~Goldberg, ``Daydreamer: World models for physical robot learning,'' in \emph{Conference on Robot Learning}.\hskip 1em plus 0.5em minus 0.4em\relax PMLR, 2023, pp. 2226--2240.

\bibitem{kingma2013auto}
D.~P. Kingma \emph{et~al.}, ``Auto-encoding variational bayes,'' \emph{arXiv preprint arXiv:1312.6114}, 2013.

\bibitem{hafner2019learning}
D.~Hafner, T.~Lillicrap, I.~Fischer, R.~Villegas, D.~Ha, H.~Lee, and J.~Davidson, ``Learning latent dynamics for planning from pixels,'' in \emph{International conference on machine learning}.\hskip 1em plus 0.5em minus 0.4em\relax PMLR, 2019, pp. 2555--2565.

\bibitem{hafner2020mastering}
D.~Hafner, T.~Lillicrap, M.~Norouzi, and J.~Ba, ``Mastering atari with discrete world models,'' \emph{arXiv preprint arXiv:2010.02193}, 2020.

\bibitem{lancaster2023modem}
P.~Lancaster, N.~Hansen, A.~Rajeswaran, and V.~Kumar, ``Modem-v2: Visuo-motor world models for real-world robot manipulation,'' \emph{arXiv preprint arXiv:2309.14236}, 2023.

\bibitem{haider2022can}
A.~Haider and H.~Hel-Or, ``What can we learn from depth camera sensor noise?'' \emph{Sensors}, vol.~22, no.~14, p. 5448, 2022.

\bibitem{lin2023end}
H.~Lin, B.~Li, X.~Chu, Q.~Dou, Y.~Liu, and K.~W.~S. Au, ``End-to-end learning of deep visuomotor policy for needle picking,'' in \emph{2023 IEEE/RSJ International Conference on Intelligent Robots and Systems (IROS)}.\hskip 1em plus 0.5em minus 0.4em\relax IEEE, 2023, pp. 8487--8494.

\bibitem{scheikl2022sim}
P.~M. Scheikl \emph{et~al.}, ``Sim-to-real transfer for visual reinforcement learning of deformable object manipulation for robot-assisted surgery,'' \emph{IEEE Robotics and Automation Letters}, vol.~8, no.~2, pp. 560--567, 2022.

\bibitem{schulman2017proximal}
J.~Schulman \emph{et~al.}, ``Proximal policy optimization algorithms,'' \emph{arXiv preprint arXiv:1707.06347}, 2017.

\bibitem{kadi2023planet}
H.~A. Kadi and K.~Terzic, ``Planet-pick: Effective cloth flattening based on latent dynamic planning,'' \emph{arXiv preprint arXiv:2303.01345}, 2023.

\bibitem{mendonca2023alan}
R.~Mendonca, S.~Bahl, and D.~Pathak, ``Alan: Autonomously exploring robotic agents in the real world,'' \emph{arXiv preprint arXiv:2302.06604}, 2023.

\bibitem{mandi2022cacti}
Z.~Mandi, H.~Bharadhwaj, V.~Moens, S.~Song, A.~Rajeswaran, and V.~Kumar, ``Cacti: A framework for scalable multi-task multi-scene visual imitation learning,'' \emph{arXiv preprint arXiv:2212.05711}, 2022.

\bibitem{hansen2022modem}
N.~Hansen, Y.~Lin, H.~Su, X.~Wang, V.~Kumar, and A.~Rajeswaran, ``Modem: Accelerating visual model-based reinforcement learning with demonstrations,'' \emph{arXiv preprint arXiv:2212.05698}, 2022.

\bibitem{mendonca2023structured}
R.~Mendonca, S.~Bahl, and D.~Pathak, ``Structured world models from human videos,'' \emph{arXiv preprint arXiv:2308.10901}, 2023.

\bibitem{yang2023track}
J.~Yang, M.~Gao, Z.~Li, S.~Gao, F.~Wang, and F.~Zheng, ``Track anything: Segment anything meets videos,'' \emph{arXiv preprint arXiv:2304.11968}, 2023.

\bibitem{allan20192017}
M.~Allan, A.~Shvets, T.~Kurmann, Z.~Zhang, R.~Duggal, Y.-H. Su, N.~Rieke, I.~Laina, N.~Kalavakonda, S.~Bodenstedt \emph{et~al.}, ``2017 robotic instrument segmentation challenge,'' \emph{arXiv preprint arXiv:1902.06426}, 2019.

\bibitem{shvets2018automatic}
A.~A. Shvets, A.~Rakhlin, A.~A. Kalinin, and V.~I. Iglovikov, ``Automatic instrument segmentation in robot-assisted surgery using deep learning,'' in \emph{2018 17th IEEE international conference on machine learning and applications (ICMLA)}.\hskip 1em plus 0.5em minus 0.4em\relax IEEE, 2018, pp. 624--628.

\bibitem{garcia2021image}
L.~C. Garcia-Peraza-Herrera, L.~Fidon, C.~D’Ettorre, D.~Stoyanov, T.~Vercauteren, and S.~Ourselin, ``Image compositing for segmentation of surgical tools without manual annotations,'' \emph{IEEE transactions on medical imaging}, vol.~40, no.~5, pp. 1450--1460, 2021.

\bibitem{kirillov2023segment}
A.~Kirillov, E.~Mintun, N.~Ravi, H.~Mao, C.~Rolland, L.~Gustafson, T.~Xiao, S.~Whitehead, A.~C. Berg, W.-Y. Lo \emph{et~al.}, ``Segment anything,'' in \emph{Proceedings of the IEEE/CVF International Conference on Computer Vision}, 2023, pp. 4015--4026.

\bibitem{cheng2022xmem}
H.~K. Cheng and A.~G. Schwing, ``Xmem: Long-term video object segmentation with an atkinson-shiffrin memory model,'' in \emph{European Conference on Computer Vision}.\hskip 1em plus 0.5em minus 0.4em\relax Springer, 2022, pp. 640--658.

\bibitem{hafner2019dream}
D.~Hafner, T.~Lillicrap, J.~Ba, and M.~Norouzi, ``Dream to control: Learning behaviors by latent imagination,'' \emph{arXiv preprint arXiv:1912.01603}, 2019.

\bibitem{horvath2022object}
D.~Horv{\'a}th, G.~Erd{\H{o}}s, Z.~Istenes, T.~Horv{\'a}th, and S.~F{\"o}ldi, ``Object detection using sim2real domain randomization for robotic applications,'' \emph{IEEE Transactions on Robotics}, vol.~39, no.~2, pp. 1225--1243, 2022.

\bibitem{kazanzides2014open}
P.~Kazanzides \emph{et~al.}, ``{An Open-Source Research Kit} for the da {Vinci}{\textregistered} {Surgical} {System},'' in \emph{IEEE/RSJ Int. Conf. Robot. and Autom.}, 2014, pp. 6434--6439.

\bibitem{schrittwieser2020mastering}
J.~Schrittwieser \emph{et~al.}, ``Mastering atari, go, chess and shogi by planning with a learned model,'' \emph{Nature}, vol. 588, no. 7839, pp. 604--609, 2020.

\bibitem{nair2022r3m}
S.~Nair, A.~Rajeswaran, V.~Kumar, C.~Finn, and A.~Gupta, ``R3m: A universal visual representation for robot manipulation,'' \emph{arXiv preprint arXiv:2203.12601}, 2022.

\end{thebibliography}

\section*{Appendix}
\begin{table}[t]
   \scriptsize
   \renewcommand{\arraystretch}{1.3}
   \caption{Hyperparameter for GAS}
   \centering
   \begin{tabular}{|c|c|c|c|}
   \Xhline{2\arrayrulewidth}
   \bf{Type} & \bf{Name} & \bf{Symbol}& \bf{Value} \\ \hline
   \multirow{5}{*}{DreamerV2} & Prefill Demonstration Steps & -& $3\times10^{3}$ \\
   & Prefill Random Steps &-& $7\times10^{3}$ \\
   & Replay Buffer Size & -& $5\times10^{5}$ \\
   & Training Steps & -& $1.8\times10^{6}$ \\ 
   & Timelimit & H& $300$ \\
   \hline
\multirow{1}{*}{Depth Estimation} 
& Empirical Size of Gripper Tip & $b_i$& $7$mm \\
\hline
\multirow{3}{*}{DSA} & 
   Square Size Ratio & $\alpha_{s}$& $0.15$ \\
   &Zoom-In Size Ratio & $\alpha_{zoom}$& $0.3$ \\
   &Image Size of Scalar Encoding & - & $6\times6$ \\
 \hline

\multirow{1}{*}{VC} 
 
   & Timesteps of Closing Clutch & $H_{clutch}$ & 6 \\
 \hline

 \multirow{6}{*}{Sparse Rewards} & 
   Successful Termination & -& $1$ \\
   &Failed Termination & -& $-0.1$ \\
   &Normal Progress & -& $-0.001$ \\
   &Abnormal Progress 1 & - & $-0.01$ \\
  &Abnormal Progress 2 & - & $-0.01$ \\
     &Abnormal Progress 3 & - & $-0.05$ \\

   \Xhline{2\arrayrulewidth}
   \end{tabular}
    {\raggedright Readers can refer to Table \ref{table:domain_random} for the hyperparameters of domain randomization of GAS \par}
   \label{table:hyperparam}
   \end{table}

\end{document}